\documentclass[10pt,twocolumn,letterpaper]{article}

\usepackage[pagenumbers]{cvpr} 

\usepackage{float}
\usepackage{newfloat}
\usepackage{listings}
\usepackage{subfiles} 
\usepackage{subcaption}
\usepackage{ragged2e} 
\usepackage{booktabs, makecell, multirow, tabularx}
\usepackage{bbding}
\usepackage{xcolor}
\usepackage{mathtools}
\usepackage{mathrsfs}
\usepackage{geometry} 
\usepackage{diagbox}

\usepackage{wrapfig}
\usepackage{floatflt}
\usepackage{listings}

\usepackage{enumitem} 
\usepackage{lipsum}
\usepackage{pifont}
\usepackage[dvipsnames]{xcolor}
\usepackage[most]{tcolorbox}
\definecolor{thinkcolor}{RGB}{227,196,144}
\definecolor{observecolor}{RGB}{153,201,227}
\definecolor{explorecolor}{RGB}{178,217,200}

\tcbset{
    common/.style={
		enhanced,
		arc=0mm,
		fonttitle=\large\bfseries,
		coltitle=black,
		attach boxed title to top left={xshift=0mm,
										yshift=-0.50mm},
		boxed title style={
			skin=enhancedfirst jigsaw,
			size=small,
			arc=5mm,
			bottom=0mm,
			left=8mm,
			right=18mm,
			top=1mm},
			boxrule=0pt,
			frame hidden},
    thinkstyle/.style={
		common,
		colbacktitle=thinkcolor,
		colframe=thinkcolor,
		colback=thinkcolor!40,
		borderline north={4pt}{0pt}{thinkcolor}},
	observestyle/.style={
		common,
		colbacktitle=observecolor,
		colframe=observecolor,
		colback=observecolor!40,
		borderline north={4pt}{0pt}{observecolor}}
}

\definecolor{interest_colframe}{rgb}{0.8, 0.878, 0.871}
\definecolor{interest_colback}{rgb}{0.918, 0.953, 0.949}

\definecolor{tag_colframe}{rgb}{0.965, 0.898, 0.847}
\definecolor{tag_colback}{rgb}{0.988, 0.961, 0.941}

\definecolor{exp_colback}{rgb}{0.949, 0.965, 0.980}
\definecolor{exp_colframe}{rgb}{0.878, 0.922, 0.965}

\tcbset{
    promptbox/.code args={#1/#2}{
        \tcbset{
            enhanced,
            arc=0mm,
            colframe=#1, 
            colback=#2, 
            coltitle=black,
            fonttitle=\large\bfseries,
            attach boxed title to top left={xshift=0mm, yshift=-1.0mm},
            boxed title style={
                skin=enhancedfirst jigsaw,
                size=small,
                arc=3mm,
                bottom=0mm,
                left=8mm,
                right=8mm,
                top=1mm,
                colback=#1
            },
            boxrule=0pt,
            frame hidden,
            borderline north={4pt}{0pt}{#1},
            float,
            floatplacement=top,
        }
    }
}

\newtcolorbox{think}{thinkstyle,title=Prompt Template}
\newtcolorbox{observe}{observestyle,title=observe}

\newtcolorbox{custom}[2][gray]{BlueViolet
	common,
	title=#2,
	colbacktitle=#1,
	colframe=#1,
	colback=#1!40,
	borderline north={4pt}{0pt}{#1}}

\definecolor{cvprblue}{rgb}{0.21,0.49,0.74}
\usepackage[pagebackref,breaklinks,colorlinks,allcolors=cvprblue]{hyperref}

\def\paperID{3905} 
\def\confName{CVPR}
\def\confYear{2026}

\newcommand{\XGraw}{\textbf{X}^{\mathcal{G}}}
\newcommand{\XTraw}{\textbf{X}^{\mathcal{T}}}
\newcommand{\XGnew}{\textbf{X}^{\mathcal{G}'}}

\newtheorem{assumption}{Assumption}[section]
\newtheorem{definition}{Definition}[section]

\newtheorem{theorem}{Theorem}[section]

\title{BLEG: LLM Functions as Powerful fMRI Graph-Enhancer \\ for Brain Network Analysis}

\author{
    Rui Dong\quad 
    Zitong Wang\quad 
    Jiaxing Li\quad 
    Weihuang Zheng\quad 
    Youyong Kong\thanks{Corresponding author}\\
    \small School of Computer Science and Engineering, Southeast University\\
    {\tt\small \{dongrui\_0427,220242336,jiaxing\_li,zhengweihuang,kongyouyong\}@seu.edu.cn}
}

\begin{document}
\maketitle
\begin{abstract}
Graph Neural Networks (GNNs) have been widely used in diverse brain network analysis tasks based on preprocessed functional magnetic resonance imaging (fMRI) data. 
However, their performances are constrained due to high feature sparsity and inherent limitations of domain knowledge within uni-modal neurographs. 
Meanwhile, large language models (LLMs) have demonstrated powerful representation capabilities. Combining LLMs with GNNs presents a promising direction for brain network analysis. While LLMs and MLLMs have emerged in neuroscience, integration of LLMs with graph-based data remains unexplored. 
In this work, we deal with these issues by incorporating LLM's powerful representation and generalization capabilities. 
Considering great cost for directly tuning LLMs, we instead \textbf{function LLM as enhancer} to boost GNN's performance on downstream tasks. 
Our method, namely \textbf{BLEG}, can be divided into three stages. We firstly prompt LLM to get augmented texts for fMRI graph data, then we design a ``LLM-LM'' instruction tuning method to get enhanced textual representations at a relatively lower cost. GNN is trained together for coarsened alignment. Finally we finetune an adapter after GNN for given downstream tasks. Alignment loss between LM and GNN logits is designed to further enhance GNN's representation.
Extensive experiments on different datasets confirmed BLEG's superiority.
Code can be available at https://github.com/KamonRiderDR/BLEG.
\end{abstract}

\begin{figure}
    \centering
    \includegraphics[width=0.95\linewidth]{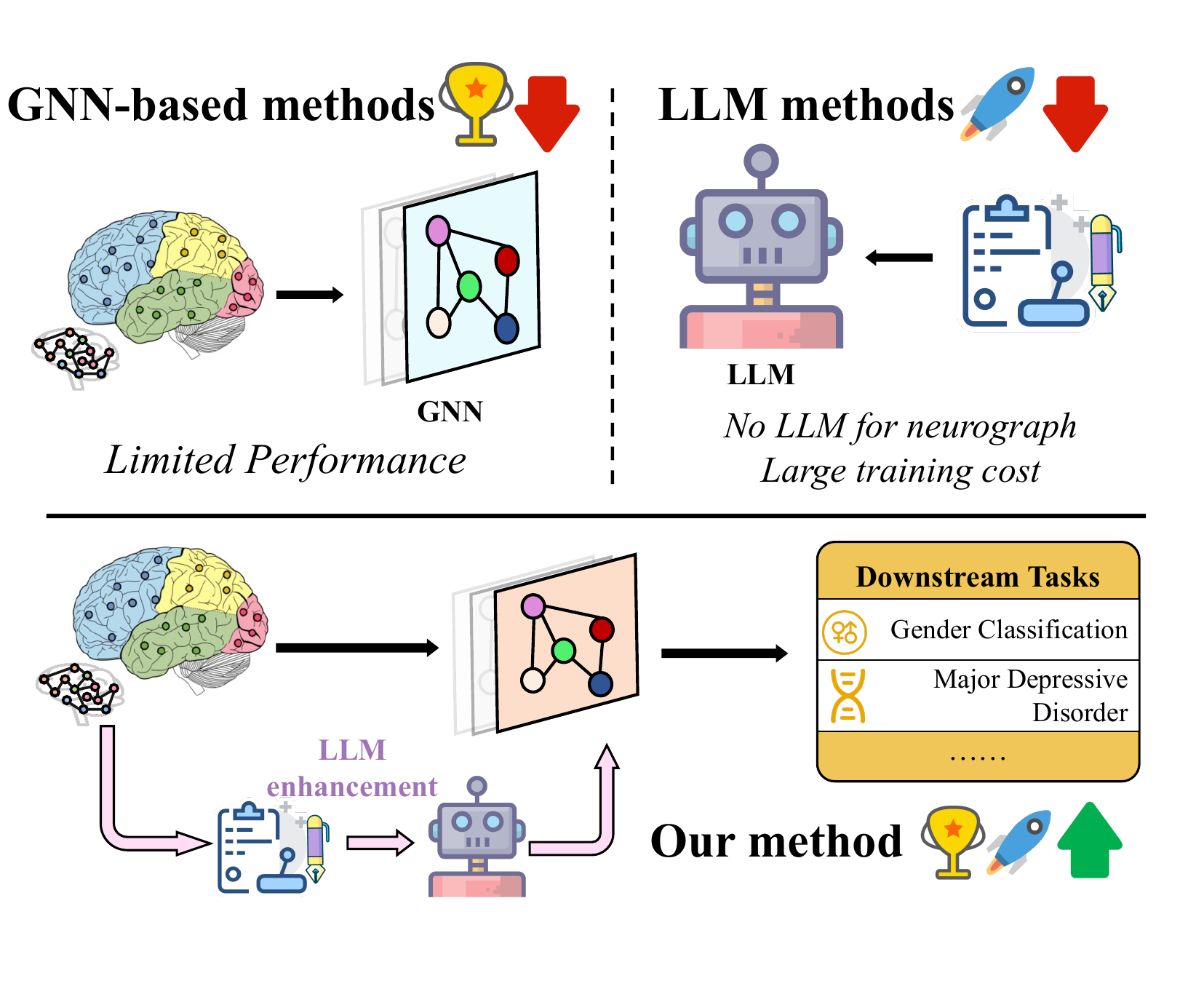}
    \caption{A illustration of our method. GNN-based methods have limited performance. LLM methods require great training cost. Our method aims to enhance GNN's performance with much less training cost.}
    \label{fig_toy}
\end{figure}
\vspace{-1em}

\section{Introduction}
\label{sec_intro}

Brain network analysis holds significant importance for investigating intrinsic mechanisms of human brains and diagnosing neurological diseases.
Deep learning-based methods have emerged as the predominant approach for brain network analysis, demonstrating superior performance for different tasks (gender classification \cite{hcp}, major depressive disorder diagnosis \cite{2024MDD}, autism spectrum disorder diagnosis \cite{2023ASD}, etc).
Among these approaches, graph neural networks (GNNs) have achieved remarkable success \cite{2017GCN, 2018GAT, 2017GraphSAGE}. By operating message passing mechanism on fMRI-derived brain networks, GNN can effectively exploit their intrinsic topological features \cite{2017fmri, 2024fmri}.
Current GNN approaches mainly focus on two directions: (a) designing powerful modules to capture finer features (DTN \cite{2016DTN}, A-GCL \cite{2023A-GCL}, BrainNPT \cite{2024BrainNPT}, etc). (b) enhancing model interpretability (BrainGNN \cite{2021BrainGNN}, IBGNN \cite{2022IBGNN}, ContrastPool \cite{2024ContrastPool}, etc).

Despite their success, performances of these methods are constrained by inherent limitations in brain graph data. Limited sample size \cite{2024lowquality} and sparse feature for preprocessed neurograph hinder further capability for data-driven deep learning methods \cite{2020sparsegraph, 2024sparse}. Meanwhile, brain network data are inherently confined to certain neuroimaging data, which lacks domain knowledge that is not explicitly encoded in imaging data. These features are further sparsified by preprocessing pipeline, leading to certain information loss; both factors jointly constrain GNN-based brain network analysis.
In short, These inherent data limitations together pose challenges for more accurate GNN-based brain network analysis.


Meanwhile, the emergence of large language models (LLMs) has achieved remarkable success in Natural Language Processing (NLP) domain, for their exceptional representation reasoning and generalization capabilities \cite{achiam2023gpt4, 2024llama3}. 
Existing LLM methods for neuroscience and brain network analysis mainly directly utilize LLMs' capabilities for different tasks which focus on single text modality (e.g. BioGPT \cite{luo2022biogpt}, BioBERT \cite{2020biobert}). They either employ multimodal language models for data from different modalities to help diagnosis \cite{chen2024medblip, 2024mmgpl}, including medical images (e.g. MedBLIP \cite{chen2024medblip}, LLaVA-Med \cite{li2023llavamed}), electrophysiological recordings (e.g. BrainBERT \cite{2023brainbert}) and structured clinical data \cite{zheng2024mmmoe, jiang2024medmoe}.
However, cases are more complex when dealing with graph data, which contains both node and structure features. Currently, exploration of LLM applications into graph-based brain network analysis remains unexplored, and a combination of LLMs and brain GNNs deserves further research. 

Hence, in this paper, we pioneer the integration of LLMs with GNN-based neuroscience tasks. Instead of using LLM as a decoder, we regard LLM as an enhancer, hoping to utilize its embeddings to enhance GNN's representation learning (shown in Fig. \ref{fig_toy}). However, directly tuning for LLM is costly. Thus, we design a novel framework which realizes \underline{\textbf{L}}anguage-\underline{\textbf{E}}nhanced \underline{\textbf{G}}raph Neural Network for \underline{\textbf{B}}rain Network Analysis (\textbf{BLEG}). Our BLEG can be divided into three stages: (1) We prompt LLM to generate augmented text description data for input graph. Each brain graph is prompted as text format. (2) We tune a smaller LM based on the text-graph dataset from previous stage. GNN encoder is also trained for coarse alignment. (3) We conduct supervised fine-tuning on GNN for different downstream tasks. Logits from tuned LM is utilized for fine-grained alignment.

Besides, we provide a theoretical analysis to demonstrate effectiveness of BLEG. By using LLM and LM as an enhancer, GNN can learn better representation beneficial for given downstream tasks. Extensive experiments on various real-world datasets to illustrate BLEG's superior performance on different tasks (ASD diagnosis, MDD diagnosis, etc). To the best of our knowledge, this is the first attempt to improve GNN brain networks' performance by exploring LLM methods. 
Notably, here only fMRI data is used while our BLEG is data-agnostic and model-agnostic, and we believe that it provides new insight for both research and real-scene applications.

\begin{figure*}[!htbp]
    \centering
    \includegraphics[width=\textwidth]{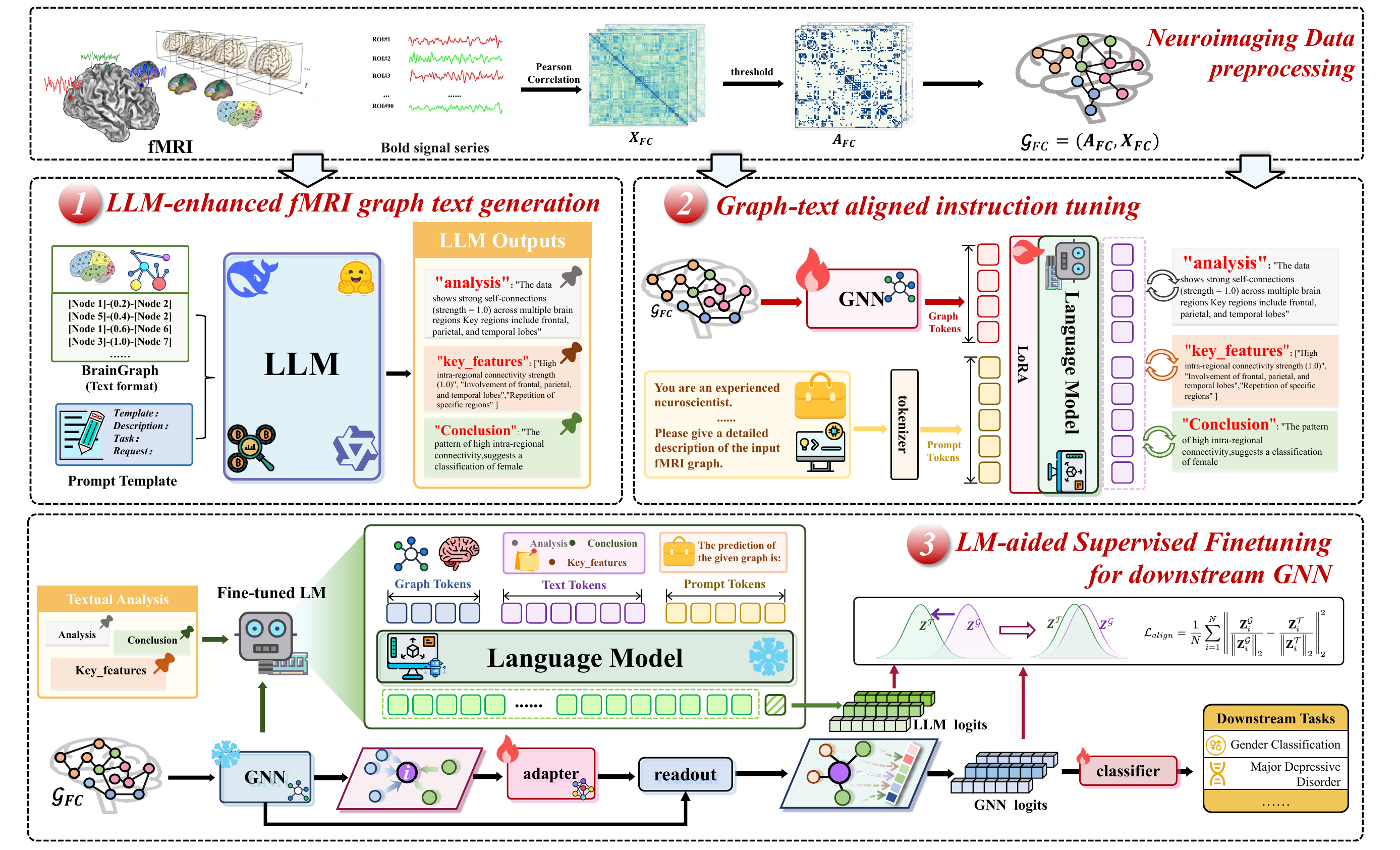}
    \caption{The overall framework of BLEG. (1) We prompt LLM to generate augmented text data for input graph. (2) A smaller LM is trained through instruction tuning based on generated textual data. GNN is trained together for coarsened alignment. (3) GNN tuning for certain tasks. Logit from LM is utilized to boost GNN's representations.}
    \label{model}
\end{figure*}



    
\section{Related Works}
\textbf{GNN-based Brain network analysis.} Graph neural networks (GNNs) follow message passing mechanism which aggregates its neighborhood nodes before updating the value of its node feature (GCN \cite{2017GCN}, GAT \cite{2018GAT}, GraphSAGE \cite{2017GraphSAGE}).
GNN-based methods in neuroscience can be mainly divided into two categories: (a) Designing modules to capture deeper features within brain graphs \cite{2016DTN, 2024thfcn}. A-GCL employs adversarial module to enhance GNN's performance \cite{2023A-GCL}. BrainNPT uses transformer as backbone to capture long-range features \cite{2024BrainNPT}. (b) Enhancing model interpretability. BrainGNN is a classical interpretable brain graph neural network \cite{2021BrainGNN}. IBGNN employs GNN backbone and generator for explanation \cite{2022IBGNN}. ContrastPool is a differentiable graph pooling method which can realize explainable classification.

\noindent\textbf{LLM methods in neuroscience.} Based on pretrained LMs (GPT, BERT), methods like BioGPT \cite{luo2022biogpt}, BioBERT \cite{2020biobert} train a medical language model from medical datasets like Pubmed. Meanwhile, the advent of multimodal LLMs (MLLMs) can utilze information from other modalities to capture complementary features. LLaVA-Med realizes a large Language-and-Vision assistant for biomedicine \cite{li2023llavamed}. MMGPL \cite{2024mmgpl} designs graph prompt for fMRI images for better text-image fusion. Other methods consider using more modalities (video \cite{saab2024gemini}, structured clinical data \cite{zheng2024mmmoe}) and new model architectures (Mixture-of-Experts, MoE) for scalability \cite{jiang2024medmoe}.

\section{Methodology}

In this section, we will discuss more details of BLEG. As is shown in Fig. \ref{model}, BLEG can be divided into three stages: (a) We firstly \textbf{conduct augmented text-graph dataset} by prompting LLM for every fMRI brain graph. (b) Then we finetune a smaller LM based on the conducted dataset through \textbf{instruction tuning}. We also train a GNN model for coarse-grained text-graph alignment. And finally (c) we perform \textbf{supervised fine-tuning} for given downstream task on pretrained GNN. We add a trainable adapter while keeping GNN weights frozen. Alignment loss is designed between GNN and LM logits for fine-grained alignment.

\subsection{Notations}
Brain graph datasets are preprocessed from functional magnetic resonance imaging (fMRI, also write as FC). Pipeline for Data preprocessing is shown in Fig. \ref{model}. AAL template is employed for FC data, and finally we denote brain network as $\mathcal{G}_i = \{X_i, A_i, \mathcal{V}_i, \mathcal{E}_i\}$, where nodes $\mathcal{V}_i$ represents different brain regions in the brain while edges $\mathcal{E}_i$ represents the connectivity between two brain regions. $X_i$ denotes node features for each region and $A_i$ denotes adjacency matrix. The dataset can be written as $\mathcal{D} = \{\mathcal{G}_i, y_i\}_{i=1}^{N}$, where $y_i \in \{0,1\}$ which stands for different tasks (gender classification, MDD diagnosis, ASD diagnosis, etc). In general, Brain network analysis can be seen as a graph classification task:
\begin{equation}
    \mathcal{F}(\mathcal{G}_i) = \text{GNN}(\mathcal{G}_i) \to \hat{y_i} 
\end{equation}

$\mathcal{F}$ can be any deep learning method and we hope to learn an optimal function to predict $y$. Here we use GNN as our $\mathcal{F}(\cdot)$, whose message passing function can be written as Eq. \ref{gnn_equ}, where $h^l_i$ is the representation for node $v_i$ in $l\mbox{-}th$ layer, \textbf{AGG}($\cdot$) stands for the aggregation of its neighborhood nodes $\mathcal{N}_i$, and \textbf{UPDATE}($\cdot$) means the update operation of $h_i$.
\begin{equation}
\begin{aligned}
h^{l+1}_i = \textbf{UPDATE}(h_i^l, \ \textbf{AGG}(\{h_j^l | j \in \mathcal{N}_i\}))
\end{aligned}
\label{gnn_equ}
\end{equation}

\subsection{LLM-enhanced text data generation}

The core idea of this stage is to leverage LLM's capability for downstream GNN. However, for neuroscience, text data is limited where data forms as graph. Meanwhile, regular manual process for creating such text data for each brain network is time-consuming and requires too much human labor. Thus, we propose to fully utilize existing LLMs to generate augmented textual information for each brain network graph.

\noindent\textbf{Prompt design.} Compared to other graphs like social networks, fMRI-based FC graphs have more medical meanings, where each node stands for specific brain regions while edge weights represent connection strengths between two brain regions. Thus, we design a prompt suited for input graph $\mathcal{G}_i$.
Formally, for given $\mathcal{G}_i$, our designed prompt consists of three parts: description $\mathcal{P}_i^{D}$, graph data ($\mathcal{P}_i^{G}$) and query ($\mathcal{P}_i^{Q}$). $\mathcal{P}_i^{D}$ depicts basic medical information from given brain network, including which dataset it belongs to, preprocess template, names of each brain region in neuroscience and possible downstream tasks. Design for prompts of input graph is challenging which may consider topological information during graph-text transformation \cite{2024graphgpt, 2024cangraphllm}. Unlike other graph learning tasks, brain networks have more concrete medical meanings, where $\mathcal{E}_{ij}$ stands for connection strength between region $i$ and region $j$. On the other hand, brain graph is sparser compared to other graph data, with a limited number of edges. Thus, for its structure prompt, we consider modifying graph as ``Node$[i]$-$x$-Node$[j]$'' format for each $\mathcal{E}_{ij}$, where $x$ means its connection strength. For node feature, we consider its mean value ($X^{d} \to X^1$). Structure prompt and feature prompt together make $\mathcal{P}_i^{G}$. For query $\mathcal{P}_i^{Q}$, we require LLM output json format data containing analysis, key features and conclusion for input FC graph.


These different prompts together make our prompt: $(\mathcal{P}_i^{D}, \mathcal{P}_i^{G}, \mathcal{P}_i^{Q}) \to \mathcal{P}_i$, which is used as LLM's input. Here we select Deepseek-v3 \cite{liu2024deepseek} as our LLM API. For each $\mathcal{G}_i$, we feed corresponding $\mathcal{P}_i$ into LLM and get response ($\mathcal{T}_i$):
\begin{equation}
    \label{equ_prompt}
    \mathcal{T}_i = \textbf{LLM}(\mathcal{P}_i)
\end{equation}

\noindent\textbf{Data verification and refinement.} For each brain network $\mathcal{G}_i$, we get its LLM-enhanced text data. To further improve quality of LLM generation, we utilize LLM (QwQ-32B \cite{qwq}) for judgment and refinement. Quality of $\mathcal{T}_i$ is measured and scored from different dimensions (professional expressions, content relevance, generation repetition). $\mathcal{T}_i$ with lower scores are then refined by LLM. Remaining generations are corrected via professional medical experts. Here that we don't focus on design of data verification process, which can be left as future work. Finally, we curate a high-quality text-enhanced graph dataset: $\mathcal{D}'_i \leftarrow (\mathcal{G}_i, \mathcal{T}_i)$.


\subsection{Graph-text aligned instruction tuning}

Directly training for LLM is extremely challenging, which requires great expense for training. Meanwhile, some LLMs like GPT4, Deepseek-v3 only provides generated text, while in some cases LLM embeddings are necessary. Here, we choose to tune a smaller language model (LM) instead of directly tuning LLM \cite{2023tape}. For enhanced graph-text data-pair $(\mathcal{G}_i, \mathcal{T}_i)$, we feed them into LM for tuning. However, there exists certain modality gap between graph and text representations on high-dimensional manifold space. Encouraged by LLaVA \cite{2023llava}, we use GNN's embeddings and textual embeddings together as LM's input. As is shown in Fig. \ref{model}, we keep both GNN encoder and LM trainable to achieve coarsened alignment between graph and text embeddings.
\begin{equation}
    \label{equ_lm}
    \begin{aligned}
        \textbf{H}_i = \textbf{LM}(\textbf{X}_i^{\mathcal{G}}, \textbf{X}_i^{\mathcal{T}})
        = \textbf{LM}(f_{\phi}(\mathcal{G}_i), \textbf{X}_i^{\mathcal{T}})
    \end{aligned}
\end{equation}

For tuning method for GNN and LM, here we use instruction tuning. LM generates answers ($\textbf{X}^{\mathcal{A}}$) for given graph data ($\textbf{X}^{\mathcal{G}}$) and question ($\textbf{X}^{\mathcal{Q}}$). 
Questions are queries requiring LM to make detailed description on given brain network. The answers are from the augmented text generated by LLM from previous stage, Generally, the input embeddings from Eq. \ref{equ_lm} can be written as follows:
\begin{equation}
    \label{equ_input}
    \textbf{X}_i = \text{concat}(f_{\phi}(\mathcal{G}_i), \textbf{X}_i^{\mathcal{Q}}, \textbf{X}_i^{\mathcal{A}})
\end{equation}

LM instruction tuning is trained through auto-regressive loss, and we only compute loss function and optimize models based on answer tokens, which can be formatted as Eq. \ref{equ_autoregressive}:
\small
\begin{equation}
    \label{equ_autoregressive}
    p(\mathbf{X}^{\mathcal{A}}|\mathbf{X}^{\mathcal{G}},\mathbf{X}^{\mathcal{Q}})=\prod_{i=1}^{L}p_{\theta}(x^{i}|\mathbf{X}^{\mathcal{G}},\mathbf{X}^{\mathcal{Q},<i},\mathbf{X}^{\mathcal{A},<i})
\end{equation}

Through graph-text aligned LM instruction tuning, we can obtain textual representations for given brain graph at a much smaller training cost.

\subsection{LM-aided finetuning for GNN}

The final stage of BLEG is LM-aided supervised finetuning for different downstream tasks. LM logit is utilized to assist downstream GNN for better representation.  
To be specific, we save weights of GNN encoder and LM after instruction tuning and keep them frozen in this stage. As is shown in Fig. \ref{model}, we add a trainable adapter after frozen GNN, which is a two-layer FFN (denote as $g_{\varphi}$) for implementation. The embeddings of graph embeddings can be formatted as:
\begin{equation}
    \label{equ_gnn_embedding}
    \mathbf{Z}_i=f_\phi\circ g_\varphi(\mathcal{G}_i)
\end{equation}

For $\mathbf{Z}_i \in \mathbb{R}^{N\times d}$, we use $\text{READOUT}(\cdot)$ function to get graph-level logits ($\mathbf{Z}_i^{\mathcal{G}}$). The READOUT function incorporates residual connection along with batch normalization (Eq. \ref{equ_gnn_logits}).

\begin{equation}
    \label{equ_gnn_logits}
    \mathbf{Z}_i^{\mathcal{G}}=\mathrm{READOUT}(\text{Norm}(\mathbf{Z}_i + \textbf{X}_i^{\mathcal{G}}))
\end{equation}

Finally $\mathbf{Z}_i^{\mathcal{G}}$ is fed into a trainable classification head for prediction, with cross-entropy loss used for model optimization ($\mathcal{L}_{CE}$).

To further enhance GNN's ability to capture text-augmented representation, we introduce an auxiliary alignment loss ($\mathcal{L}_{align}$) between text ($\mathbf{Z}_i^{\mathcal{T}}$) and graph logits ($\mathbf{Z}_i^{\mathcal{G}}$). $\mathbf{Z}_i^{\mathcal{T}}$ is obtained through tuned LM, the input is the same format as Eq. \ref{equ_lm}, where $\textbf{X}_i^{\mathcal Q}$ is about give the prediction result of the input brain network. We add a \verb|cls| token at the end of each input sequence whose output logit is used as $\mathbf{Z}_i^{\mathcal{T}}$ for fine-grained graph-text alignment. For implementation of $\mathcal{L}_{align}$, we use $\text{MSE}(\cdot)$ for alignment at high manifold dimension (Eq. \ref{equ_align}).
\begin{equation}
    \label{equ_align}
    \mathcal{L}_{align} =
    \frac{1}{N}\sum_{i=1}^{N} 
    \left\| 
    \frac{\textbf{Z}_i^{\mathcal{G}}}{\left\| \textbf{Z}_i^{\mathcal{G}} \right\|_2} 
    - 
    \frac{\textbf{Z}_i^{\mathcal{T}}}{\left\| \textbf{Z}_i^{\mathcal{T}} \right\|_2} 
    \right\|_2^2
\end{equation}

The overall loss function is composed of $\mathcal{L}_{CE}$ and $\mathcal{L}_{align}$, weighted by coefficient $\alpha \in (0, 1)$:

\begin{equation}
    \label{equ_loss}
    \mathcal{L} = \mathcal{L}_{CE} + \alpha\cdot \mathcal{L}_{align}
\end{equation}
\begin{table*}[!htbp]
    \centering

    \caption{Comparison results on public datasets. We run 10 times for each model and record the corresponding average acc $\pm$ std (\%). The best results are marked \textbf{bold}, and second best \underline{underline}.}
    \label{table_comp_public}

    \resizebox{0.92\textwidth}{!}{
    \begin{tabular}{l|cc|cc|cc|cc}
        \toprule
        \multirow{2}{*}{\large{\textbf{Methods}}} &
        \multicolumn{2}{c|}{\textbf{HCP}} &
        \multicolumn{2}{c|}{\textbf{ADHD}} &
        \multicolumn{2}{c|}{\textbf{MDD}} &
        \multicolumn{2}{c}{\textbf{ABIDE}}
        \\
        &

        \small{ACC} & \small{SEN} &
        \small{ACC} & \small{SEN} &
        \small{ACC} & \small{SEN} &
        \small{ACC} & \small{SEN} 
        \\
        \midrule
        \textbf{GCN} &
        64.03 $\pm$ \scriptsize{1.21} & 55.51 $\pm$ \scriptsize{2.88} &
        66.74 $\pm$ \scriptsize{1.47} & 31.83 $\pm$ \scriptsize{2.33} &
        62.38 $\pm$ \scriptsize{0.37} & \underline{70.88 $\pm$ \scriptsize{1.94}} &
        69.14 $\pm$ \scriptsize{0.84} & 64.07 $\pm$ \scriptsize{2.20} 
        \\
        \textbf{GAT} &
        65.72 $\pm$ \scriptsize{0.67} & 60.91 $\pm$ \scriptsize{3.26} &
        66.56 $\pm$ \scriptsize{0.30} & 32.31 $\pm$ \scriptsize{1.56} &
        63.05 $\pm$ \scriptsize{0.35} & 69.30 $\pm$ \scriptsize{2.16} &
        68.79 $\pm$ \scriptsize{0.74} & 63.83 $\pm$ \scriptsize{3.23} 
        \\
        \textbf{GraphSAGE} &
        66.87 $\pm$ \scriptsize{0.32} & 61.32 $\pm$ \scriptsize{2.71} &
        67.80 $\pm$ \scriptsize{0.37} & 31.52 $\pm$ \scriptsize{2.54} &
        63.29 $\pm$ \scriptsize{0.27} & 69.48 $\pm$ \scriptsize{1.53} &
        \underline{70.74 $\pm$ \scriptsize{0.89}} & 65.52 $\pm$ \scriptsize{2.81} 
        \\
        \textbf{GraphTrans} &
        67.46 $\pm$ \scriptsize{0.55} & 62.71 $\pm$ \scriptsize{3.17} &
        67.00 $\pm$ \scriptsize{0.37} & 31.85 $\pm$ \scriptsize{2.54} &
        63.37 $\pm$ \scriptsize{0.27} & 68.31 $\pm$ \scriptsize{1.93} &
        68.41 $\pm$ \scriptsize{1.20} & 62.79 $\pm$ \scriptsize{2.75} 
        \\
        \midrule
        \textbf{BrainGNN} &
        66.46 $\pm$ \scriptsize{2.12} & 62.92 $\pm$ \scriptsize{2.45} &
        67.16 $\pm$ \scriptsize{2.01} & 32.75 $\pm$ \scriptsize{3.31} &
        63.25 $\pm$ \scriptsize{1.06} & 68.91 $\pm$ \scriptsize{2.76} &
        70.03 $\pm$ \scriptsize{1.69} & 62.80 $\pm$ \scriptsize{4.19} 
        \\
        \textbf{IBGNN} &
        64.72 $\pm$ \scriptsize{1.04} & 55.98 $\pm$ \scriptsize{4.01} &
        65.59 $\pm$ \scriptsize{0.49} & 30.84 $\pm$ \scriptsize{2.92} &
        63.07 $\pm$ \scriptsize{0.29} & 68.38 $\pm$ \scriptsize{2.47} &
        66.02 $\pm$ \scriptsize{1.18} & 61.31 $\pm$ \scriptsize{3.61} 
        \\
        \textbf{BrainNPT} &
        67.78 $\pm$ \scriptsize{1.53} & 60.67 $\pm$ \scriptsize{1.34} &
        67.84 $\pm$ \scriptsize{2.11} & 32.18 $\pm$ \scriptsize{2.19} &
        63.81 $\pm$ \scriptsize{0.77} & 69.55 $\pm$ \scriptsize{3.07} &
        68.80 $\pm$ \scriptsize{1.10} & 59.52 $\pm$ \scriptsize{3.05} 
        \\
        \textbf{THFCN} &
        67.29 $\pm$ \scriptsize{1.74} & 59.82 $\pm$ \scriptsize{2.84} &
        66.73 $\pm$ \scriptsize{1.40} & 33.01 $\pm$ \scriptsize{3.69} &
        62.41 $\pm$ \scriptsize{0.67} & 66.59 $\pm$ \scriptsize{2.17} &
        66.93 $\pm$ \scriptsize{1.29} & 62.37 $\pm$ \scriptsize{2.27}
        \\
        \textbf{ContrastPool} &
        68.10 $\pm$ \scriptsize{1.74} & 63.59 $\pm$ \scriptsize{1.21} &
        65.10 $\pm$ \scriptsize{0.82} & 35.92 $\pm$ \scriptsize{4.21} &
        64.05 $\pm$ \scriptsize{0.47} & 66.22 $\pm$ \scriptsize{3.68} &
        69.89 $\pm$ \scriptsize{0.88} & \underline{66.71 $\pm$ \scriptsize{2.79}}
        \\
        \midrule
        \textbf{TAPE} & 
        69.32 $\pm$ \scriptsize{1.41} & \underline{63.80 $\pm$ \scriptsize{2.93}} &
        67.41 $\pm$ \scriptsize{1.06} & 36.59 $\pm$ \scriptsize{2.11} & 
        \underline{64.12 $\pm$ \scriptsize{0.82}} & 68.50 $\pm$ \scriptsize{2.79} & 
        70.43 $\pm$ \scriptsize{0.95} & 66.64 $\pm$ \scriptsize{2.95} 
        \\
        \textbf{OFA} & 
        \underline{71.00 $\pm$ \scriptsize{0.82}} & 63.04 $\pm$ \scriptsize{2.42} & 
        \underline{68.22 $\pm$ \scriptsize{0.70}} & \underline{39.72 $\pm$ \scriptsize{1.94}} & 
        62.97 $\pm$ \scriptsize{1.21} & 68.55 $\pm$ \scriptsize{2.40} &
        69.72 $\pm$ \scriptsize{1.48} & 65.93 $\pm$ \scriptsize{3.11}
        
        \\
        \midrule
        \textbf{BLEG (Ours)} &
        \textbf{71.21 $\pm$ \scriptsize{0.91}} & \textbf{68.27 $\pm$ \scriptsize{3.39}} &
        \textbf{69.41 $\pm$ \scriptsize{0.53}} & \textbf{41.11 $\pm$ \scriptsize{2.10}} &
        \textbf{65.63 $\pm$ \scriptsize{0.48}} & \textbf{71.12 $\pm$ \scriptsize{2.97}} &
        \textbf{72.21 $\pm$ \scriptsize{0.80}} & \textbf{67.16 $\pm$ \scriptsize{2.30}} 
        \\
        \textbf{$\Delta$ GCN} & 
        \textbf{7.18} \color{red}{$\uparrow$} & \textbf{12.76} \color{red}{$\uparrow$} &
        \textbf{2.67} \color{red}{$\uparrow$} & \textbf{9.28} \color{red}{$\uparrow$} &
        \textbf{3.25} \color{red}{$\uparrow$} & \textbf{0.24} \color{red}{$\uparrow$} &
        \textbf{3.07} \color{red}{$\uparrow$} & \textbf{3.09} \color{red}{$\uparrow$} 
        \\        
        
        \bottomrule
    \end{tabular}    
    } 
\end{table*}    
\begin{table*}[!tbp]
    \centering
    \caption{Comparison results on private dataset. We run 10 times for each model and record the corresponding average acc $\pm$ std (\%). The best results are marked \textbf{bold}, and second best \underline{underline}.}
    \label{table_comp_private}

    \resizebox{0.92\textwidth}{!}{
    \begin{tabular}{l|ccccc|ccc}
        \toprule

        \textbf{Methods} &
        
        \textbf{GCN} & 
        \textbf{GraphTrans} &
        
        \textbf{BrainGNN} & \textbf{BrainNPT} & \textbf{ContrastPool} & 
        
        \textbf{BLEG (medium)} &
        \textbf{BLEG (large)} &
        \textbf{Qwen3-8B}

        \\
        \midrule
        \textbf{ACC} &
        71.79 $\pm$ \scriptsize{1.07} & 
        70.50 $\pm$ \scriptsize{0.73} &
        71.06 $\pm$ \scriptsize{1.05} & 
        70.25 $\pm$ \scriptsize{1.55} &
        71.11 $\pm$ \scriptsize{0.73} &  
        \underline{75.38 $\pm$ \scriptsize{0.63}} &  
        75.21 $\pm$ \scriptsize{1.01} & 
        \textbf{75.82 $\pm$ \scriptsize{0.77}}         
        \\

        \textbf{SEN} &
        85.70 $\pm$ \scriptsize{2.46} & 
        86.07 $\pm$ \scriptsize{1.84} &
        84.00 $\pm$ \scriptsize{1.81} & 
        83.00 $\pm$ \scriptsize{1.59} &
        86.85 $\pm$ \scriptsize{1.94} &  
        \textbf{87.03 $\pm$ \scriptsize{0.89}} &  
        86.77 $\pm$ \scriptsize{1.27} & 
        \textbf{87.03 $\pm$ \scriptsize{1.58}}         
        \\

        \textbf{SPE} &
        50.60 $\pm$ \scriptsize{5.41} & 
        50.92 $\pm$ \scriptsize{3.03} &
        48.77 $\pm$ \scriptsize{2.12} & 
        49.21 $\pm$ \scriptsize{3.30} &
        52.55 $\pm$ \scriptsize{3.91} &  
        58.31 $\pm$ \scriptsize{2.72} &  
        \underline{58.50 $\pm$ \scriptsize{1.93}} & 
        \textbf{58.81 $\pm$ \scriptsize{2.00}}         
        \\

        \textbf{F1} &
        78.50 $\pm$ \scriptsize{0.67} & 
        78.43 $\pm$ \scriptsize{0.49} &
        77.18 $\pm$ \scriptsize{1.92} & 
        77.27 $\pm$ \scriptsize{0.74} &
        78.43 $\pm$ \scriptsize{0.49} &  
        79.35 $\pm$ \scriptsize{1.66} &  
        \textbf{80.78 $\pm$ \scriptsize{1.72}} & 
        \underline{79.81 $\pm$ \scriptsize{1.58}}         
        \\

        \textbf{AUC} &
        68.15 $\pm$ \scriptsize{1.71} & 
        65.70 $\pm$ \scriptsize{1.16} &
        66.39 $\pm$ \scriptsize{0.61} & 
        67.53 $\pm$ \scriptsize{1.03} &
        68.70 $\pm$ \scriptsize{1.16} &  
        \underline{70.77 $\pm$ \scriptsize{1.16}} &  
        \textbf{70.78 $\pm$ \scriptsize{1.24}} & 
        69.43 $\pm$ \scriptsize{2.03}         
        \\
        
        \bottomrule
    \end{tabular}    
    } 
\end{table*}    

\section{Theoretical Analysis}
\label{theory}
The main idea of theoretical analysis is that through LLM and LM as enhancer, the performance of GNN will be improved for given downstream tasks as its representations contains augmented text information which is useful for downstream classification.

\begin{theorem} \textbf{(Complementary Representations from LM for GNN)}
    \label{theory_1}
    We define representation from original GNN ($\textbf{X}^{\mathcal{G}}$) and fine-tuned LM ($\textbf{X}^{\mathcal{T}}$), LM-distilled GNN representation is denoted as $\textbf{X}^{\mathcal{G'}}$. Downstream label representation is denoted as Y. Given above assumptions, we have: $\left\| I(\textbf{X}^{\mathcal{G'}};Y) -         I(\textbf{X}^{\mathcal{G}},\textbf{X}^{\mathcal{T}};Y)\right\| \le C \cdot \epsilon$, where $C$ is a constant and $\epsilon > 0$. Thus for $I(\textbf{X}^{\mathcal{G'}};Y)$ and $I(\textbf{X}^{\mathcal{G}};Y)$, $I(\textbf{X}^{\mathcal{G'}};Y) >  I(\textbf{X}^{\mathcal{G}};Y)$.

\end{theorem}

Proof can be found in Appendix. Theorem. \ref{theory_1} shows that through BLEG, GNN can capture complementary information from LM, which enhances its capability for downstream tasks.
\section{Experiments}
\label{sec_exp}



\subsection{Experimental settings}
\textbf{Datasets.} Our experiments were performed on four public real-world brain network datasets: Autism Brain Imaging Data Exchange (\textbf{ABIDE}, 618 subjects) \cite{2014ABIDE}, Human Connectsome Project (\textbf{HCP}, 1039 subjects) \cite{hcp}, Attention Deficit Hyperactivity Disorder (\textbf{ADHD}, 938 subjects) \cite{adhd} and Rest-meta-MDD (\textbf{MDD}, 2165 subjects) dataset. We also use one private dataset \textbf{zhongdaxinxiang} (short as \textbf{ZDXX}, 520 subjects), which is collected from Zhongda Hospital of Southeast University, the Second Affiliated Hospital of Xinxiang Medical University and Hangzhou Hospital. Due to data privacy concerns, in first and second stage, we only use public datasets for text generation and LM instruction tuning. More details of the datasets can be found in Appendix.



\noindent\textbf{Baselines.} We select nine representative baselines for comparison, which can mainly be divided into two types: \textbf{(a) GNNs based methods}, including GCN \cite{2017GCN}, GAT \cite{2018GAT}, GraphSAGE \cite{2017GraphSAGE} and GraphTrans \cite{wu2021representing}. \textbf{(b) Brain Network based methods}, including classical methods (BrainGNN \cite{2021BrainGNN} and IBGNN \cite{2022IBGNN}) and latest brain network methods (BrainNPT \cite{2024BrainNPT}, THFCN \cite{2024thfcn} and ContrastPool \cite{2024ContrastPool}). \textbf{(c) LLM-GNN methods}: TAPE \cite{2023tape} and OFA \cite{ofa}. Code implementations of all methods are taken from their original papers.

\noindent\textbf{Experimental settings.} For LLM, we select Deepseek-v3 to generate augmented text data. We select BioGPT-base as our instruction tuning LM, whose parameter is 347M in total. For GNN encoder. We choose a 3-layer GCN. Our model is implemented in PYG and trained on RTX Titans with 24GB memory. For instruction tuning, we tune our LM and GNN on four public datasets whose number of total sample is 4760 and tuning epoch is from 3 to 5. For supervised fine-tuning, the total training epoch is 150 with 50 as early stopping. More details of our model can refer to Appendix.

We evaluate models' performance on five metrics: accuracy (\textbf{ACC}), sensitivity (\textbf{SEN}), specificity (\textbf{SPE}), f1 score (\textbf{F1}) and ROC-AUC (\textbf{AUC}), where higher value means better performance. We record ACC and SEN for public datasets, while all five metrics on zhongdaxinxiang. For evaluation on all the methods, we use 10-fold cross validation on ten random runs and record mean value and standard deviation.

\subsection{Comparison results}

\textbf{Comparison results on public datasets.} Comparison results on public four datasets are presented in Tab. \ref{table_comp_public}. The results show that our BLEG outperforms all other methods on all the datasets. Compared to GCN which also works as our backbone, the maximum improvement of ACC can be $7.18\uparrow$ on HCP dataset. It also exceeds SOTA brain network analysis method (ContrastPool) at $3.11\uparrow$ on ACC.

\noindent\textbf{Comparison results on private dataset.} Comparison results on private dataset zhongdaxinxiang are presented in Tab. \ref{table_comp_private}. Here we select two additional LMs with larger parameters for tuning: BioGPT-1.5B and Qwen3-8B. We employ Low-Rank Adaptation (LoRA) during training \cite{lora}. BLEG also achieves best results on all evaluation metrics, although we did not use it for augmented-text generation and instruction tuning. Its maximum accuracy improvement over vanilla GCN reaches $4.03\uparrow$.

\subsection{Few-shot experiment results}
LLMs have shown remarkable performance in few-shot learning settings. Thus, to further illustrate capability of BLEG, we construct few-shot splitting of datasets and test BLEG's performance in few-shot cases.

We first set a train ratio to gradually decrease number of training samples for given dataset.
As is shown in Fig. \ref{fig_few_shot}, we set training ratio from $10\%$ to $70\%$. We set a fixed validation size ($10\%$) and the rest data is for test. Results from Fig. \ref{fig_few_shot} show that compared to other methods (GNN-based BrainGNN and transformer-based BrainNPT), BLEG can always show a leading advantage.

We also conduct $k$-shot experiments on BLEG. For given dataset, we randomly select $k$ samples from each label as training set ($k \in \{1, 2, 5\}$). For testing set, we select 50, 100 and rest data ($L_\mathcal{D}$) for each label respectively. As is shown in Tab. \ref{tab_k_way}, BLEG outperforms other methods under extreme few-shot cases. The results demonstrate BLEG's superiority under few-shot scenes.

\subsection{Ablation studies \& Sensitivity analyses}
\begin{figure}[H]
    \centering
    \includegraphics[width=0.49\linewidth]{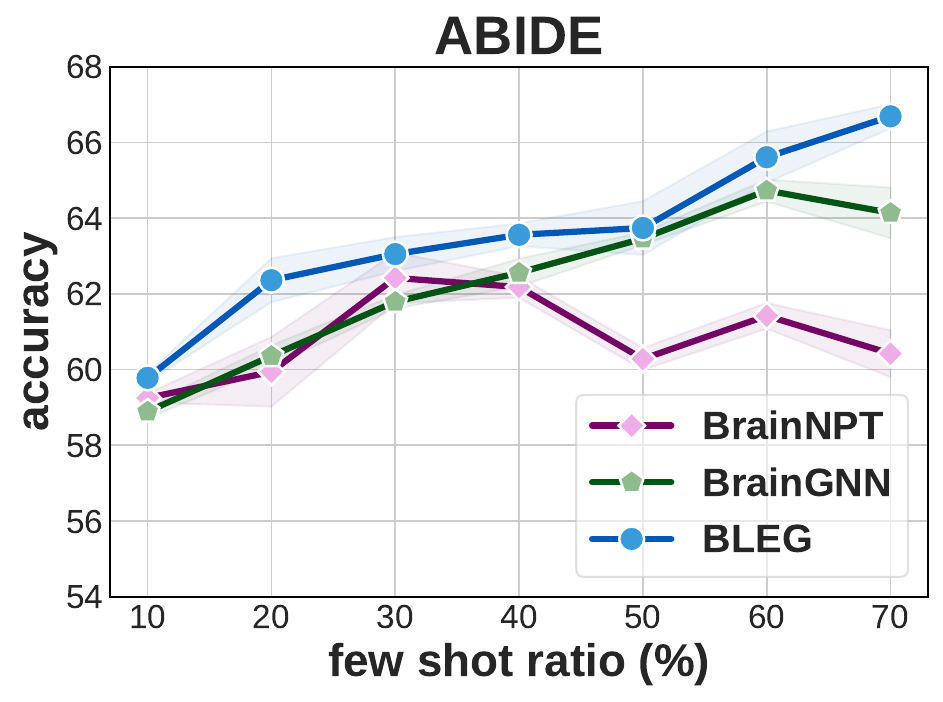}
    \includegraphics[width=0.49\linewidth]{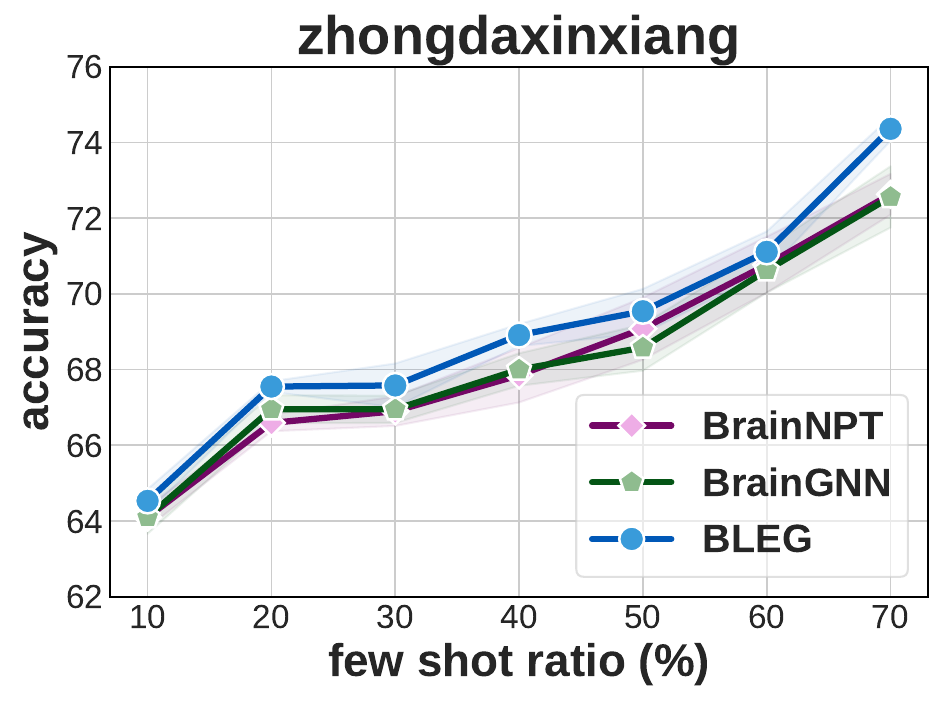}
    \caption{Few shot results on different ratios.
    }
    \label{fig_few_shot}
\end{figure}

We conduct ablation studies to analyze whether each sub-module of BLEG works. We directly train our GCN on downstream tasks to verify effectiveness of $\mathcal{L}_{align}$. We also test performance for vanilla BioGPT to verify if instruction tuning stage works. The results in Fig. \ref{fig_few_shot} (f) shows that \textbf{w/o align loss}, accuracy of BLEG will witness a great decrease. Meanwhile, $\mathcal{L}_{align}$ on BioGPT \textbf{w/o tuning} will also influence model's performance.

For sensitivity analyses, we record BLEG's accuracy under different align loss coefficient ($\alpha \in [0.2, 0.7]$) and plot following figures (Fig. \ref{fig_few_shot} (a) - (b)). The results demonstrate that the optimal value of $\alpha$ for achieving highest accuracy varies across different datasets. On the other hand, despite variations in different $\alpha$, BLEG consistently outperforms other methods (BrainGNN, BrainNPT) in most cases, further demonstrating the effectiveness of LM-aided representation enhancement.

\vspace{-2.0em}
\begin{flushleft}
\begin{table}[!tbp]
    \centering
    \caption{$k$-shot few-shot experiments. We run ten times and record average accuracy.}
    \resizebox{0.98\linewidth}{!}{
    \begin{tabular}{l|c|ccc|ccc}
        \toprule
        \multirow{2}{*}{\textbf{Methods}} &
        \multirow{2}{*}{\diagbox[dir=SE]{$N_{test}$}{$k$}} &
        \multicolumn{3}{c|}{\textbf{HCP}} & 
        \multicolumn{3}{c}{\textbf{zhongdaxinxiang}}
        \\\
        
         & &  
         \multicolumn{1}{c}{1}  &
         2 &  
         \multicolumn{1}{c|}{5} &
         \multicolumn{1}{c}{1}  &
         2 &  
         \multicolumn{1}{c}{5} 
        \\

        \midrule
        \multirow{3}{*}{\textbf{BrainGNN}} 
        & 50 &
        55.00 & 57.00 & 56.30 & 
        52.00 & 49.00 & 53.00 
        \\

        & 100 & 
        51.00 & 54.50 & 57.00 & 
        52.00 & 48.50 & 50.00 
        \\

        & $L_\mathcal{D}$ & 
        47.88 & 48.75 & 50.81 & 
        45.56 & 55.62 & 53.43 
        \\

        \midrule
        \multirow{3}{*}{\textbf{BrainNPT}} 
        & 50 &
        57.00 & 56.00 & 59.00 & 
        53.00 & 52.50 & 54.50 
        \\

        & 100 & 
        56.00 & 52.50 & 55.50 & 
        52.40 & 51.00 & 50.00 
        \\

        & $L_\mathcal{D}$ & 
        56.41 & 56.40 & 57.46 & 
        48.76 & 55.63 & 55.13 
        \\

        \midrule
        \multirow{3}{*}{\textbf{BLEG}} 
        & 50 &
        \textbf{61.00} & \textbf{59.00} & \textbf{60.30} & 
        \textbf{58.00} & \textbf{58.50} & \textbf{60.00} 
        \\

        & 100 & 
        \textbf{58.50} & \textbf{57.50} & \textbf{59.00} & 
        \textbf{54.50} & \textbf{54.00} & \textbf{56.00} 
        \\

        & $L_\mathcal{D}$ & 
        \textbf{56.81} & \textbf{56.64} & \textbf{57.53} & 
        \textbf{52.23} & \textbf{56.01} & \textbf{56.52} 
        \\

        \bottomrule
    \end{tabular}    
    } 

    \label{tab_k_way}
\end{table}
    
\end{flushleft}

\vspace{-1.0em}

\begin{figure*}[!htbp]
  \centering


  \begin{subfigure}[b]{0.3\linewidth}
    \includegraphics[width=\linewidth]{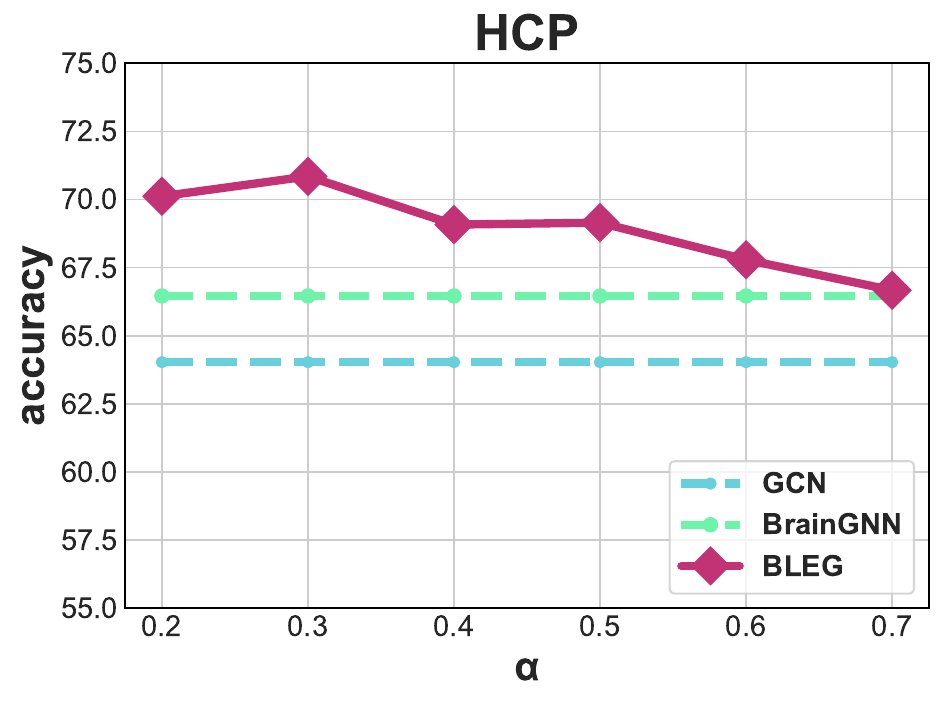}
    \caption{}\label{hyper_HCP}
  \end{subfigure}
  \begin{subfigure}[b]{0.3\linewidth}
    \includegraphics[width=\linewidth]{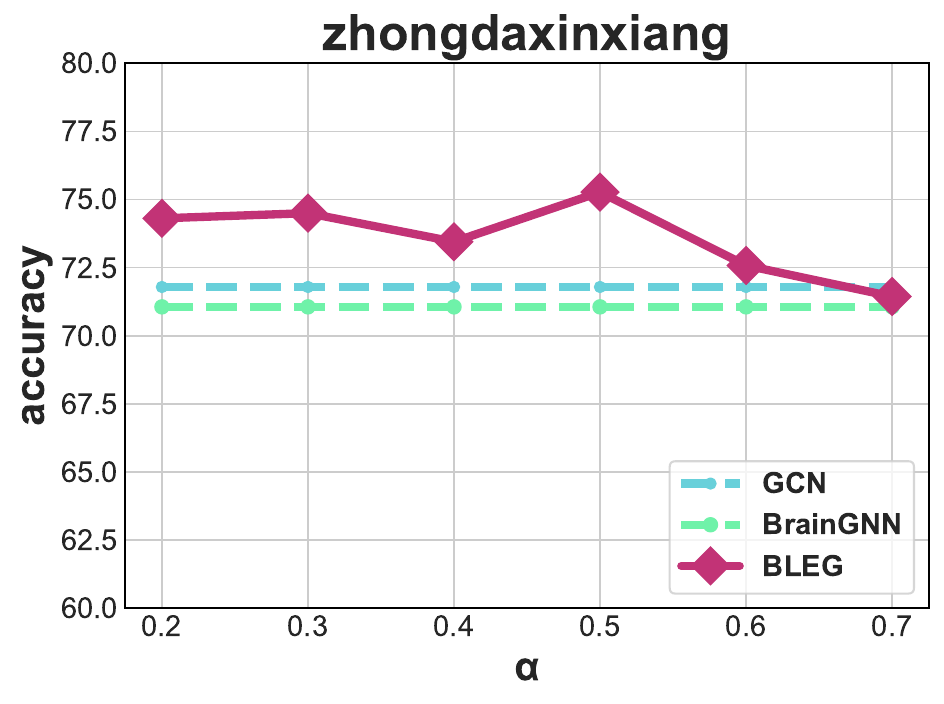}
    \caption{}\label{hyper_ZDXX}
  \end{subfigure}
  \begin{subfigure}[b]{0.3\linewidth}
    \includegraphics[width=\linewidth]{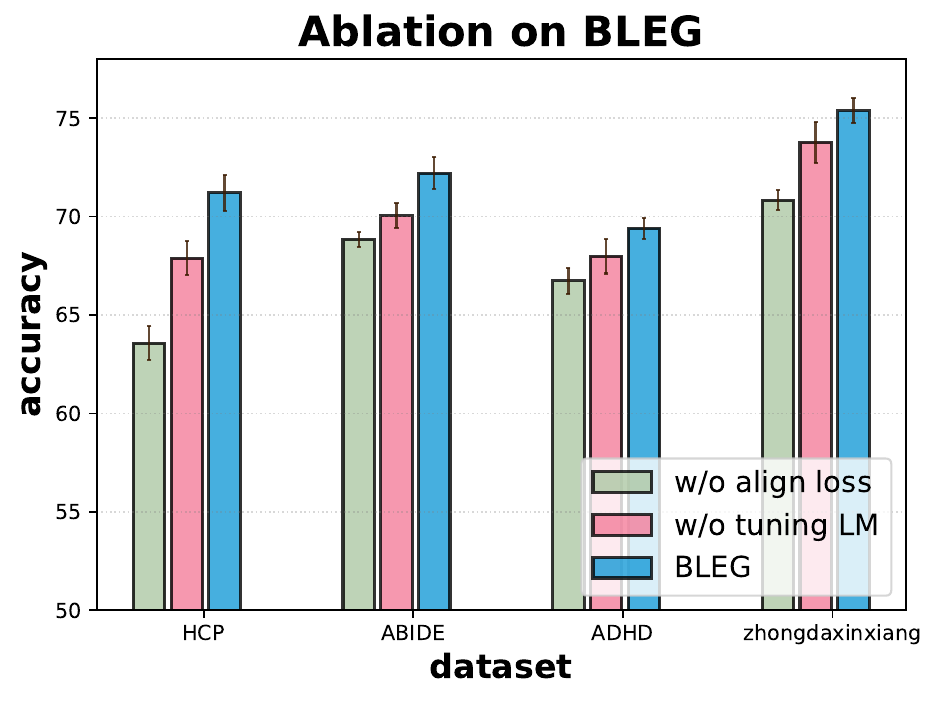}
    \caption{}\label{ablation}
  \end{subfigure}

  \begin{subfigure}[b]{0.3\linewidth}
    \includegraphics[width=\linewidth]{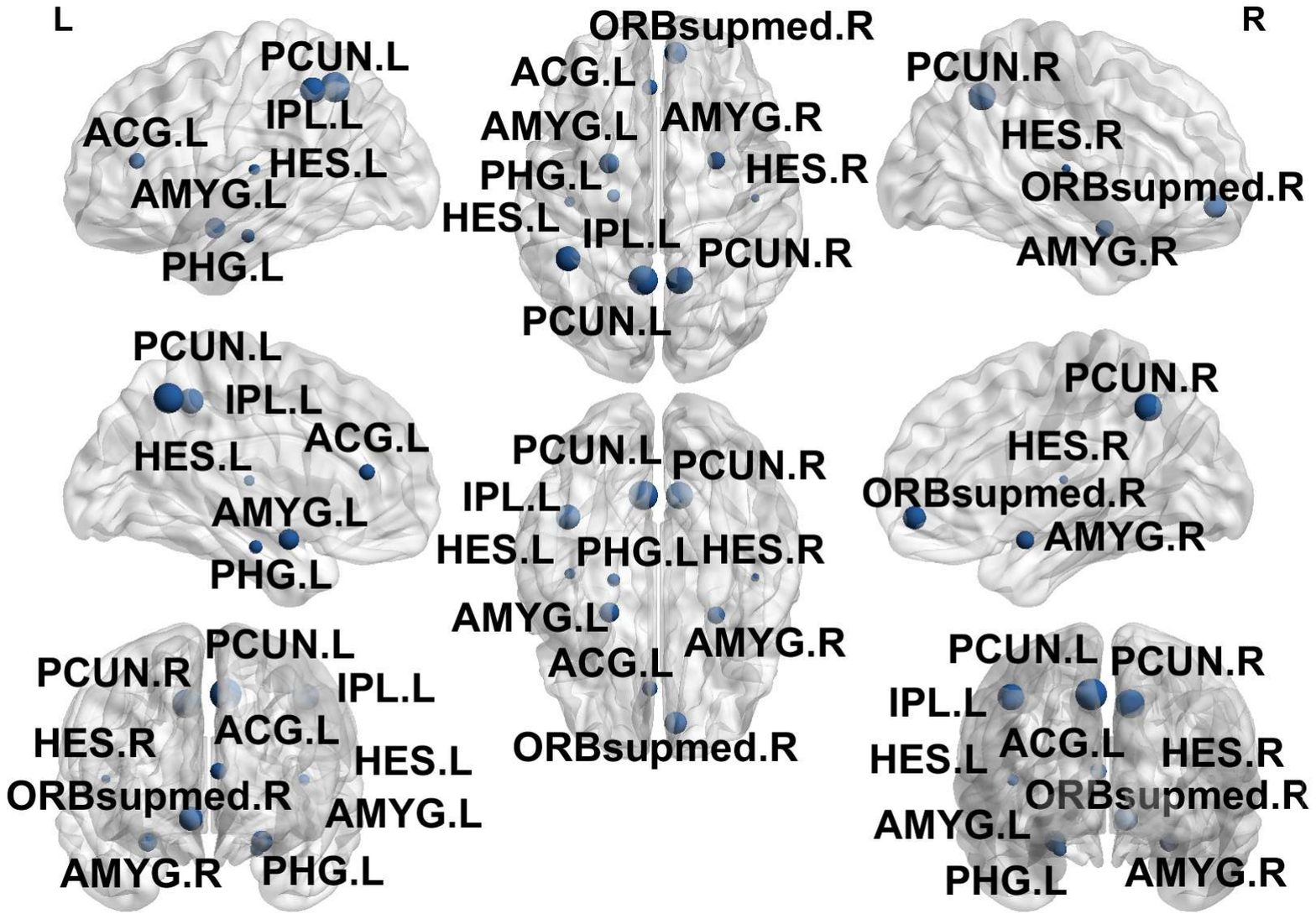}
    \caption{}\label{vis_ABIDE}
  \end{subfigure}
  \begin{subfigure}[b]{0.3\linewidth}
    \includegraphics[width=\linewidth]{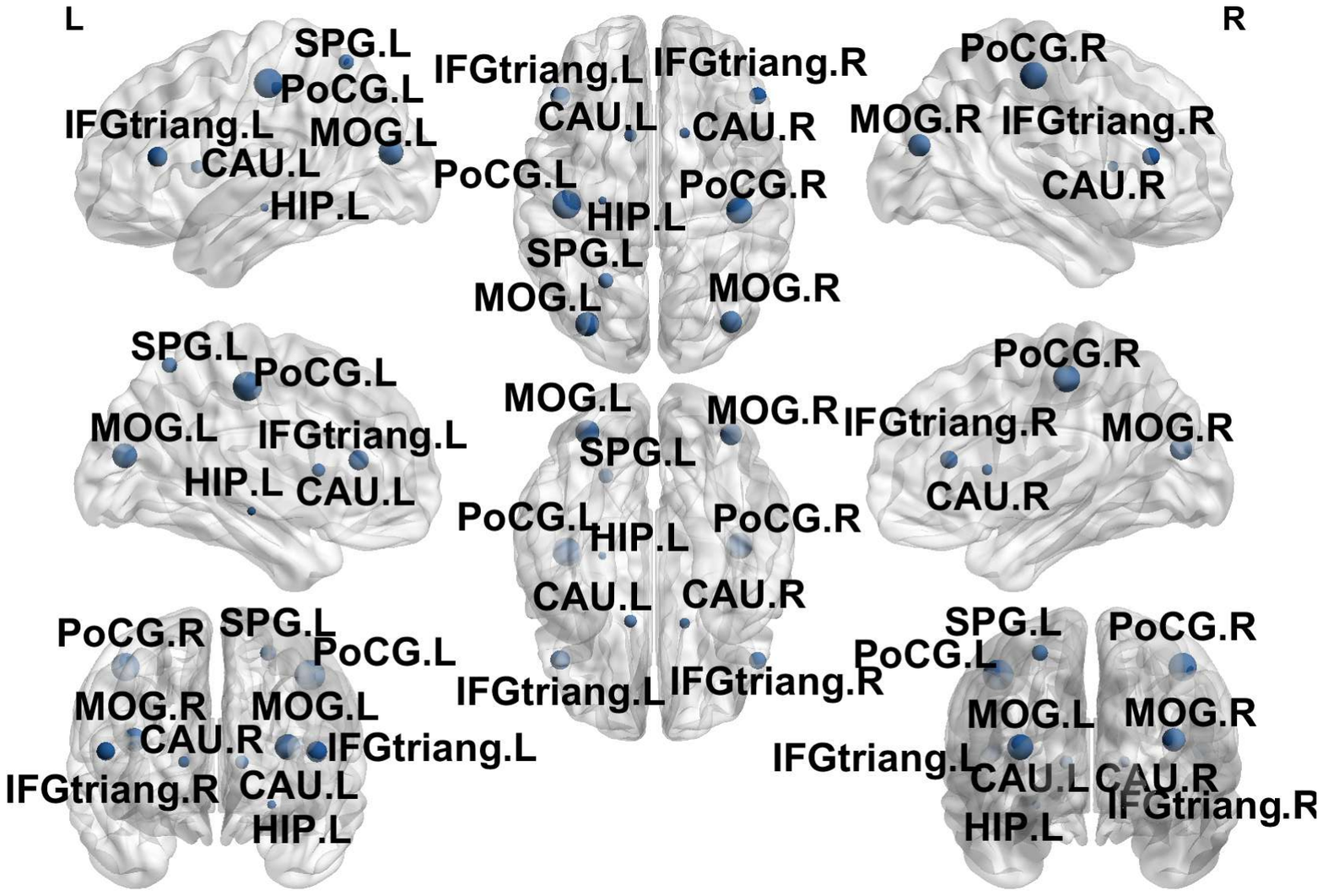}
    \caption{}\label{vis_ADHD}
  \end{subfigure}
  \begin{subfigure}[b]{0.3\linewidth}
    \includegraphics[width=\linewidth]{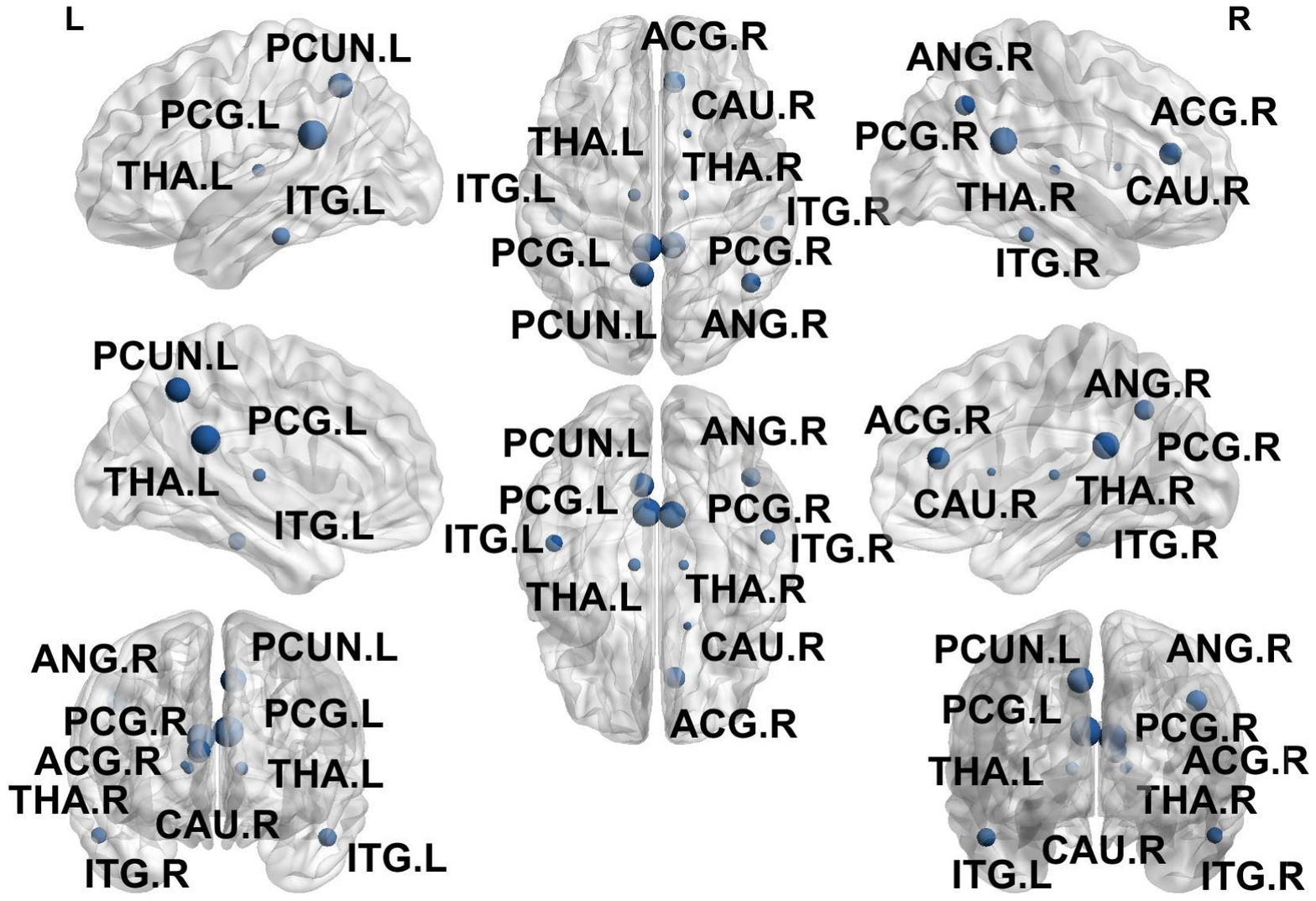}
    \caption{}\label{vis_ZDXX}
  \end{subfigure}

  \caption{(a)--(b) $k$-shot experiments on different datasets. (c) Ablation studies on different datasets. (d) Biomarker visualizations on ABIDE, (e) ADHD and (f) zhongdaxinxiang datasets.}
  \label{fig_exp}
\end{figure*}

\subsection{Empirical studies}
We conduct biomarker visualization of brain regions for empirical studies. We average embeddings after GNN encoder for all samples and statistically analyze and visualize the top 10 brain regions. The results are shown in Fig. \ref{fig_exp}. 
Visualizations for top 10 regions in ABIDE dataset are: Precuneus (L), Precuneus (R), Inferior parietal, but supramarginal and angular gyri, which are consistent with findings in \cite{2017ABIDE1}, Superior frontal gyrus and medial orbital which show additional abnormalities between HC and ASD patients \cite{2019ABIDE2}, Amygdala (L) and Amygdala (R) whose atypical activation can occur between patients \cite{2024ABIDE3}, Heschl gyrus which is positively related to ASD symptoms \cite{2022ABIDE4}, Anterior cingulate, paracingulate gyri and Parahippocampal gyrus, which are consistent with studies in \cite{2021ABIDE5}. For ZDXX dataset, brain regions like Posterior cingulate gyrus, Precuneus, Anterior cingulate and paracingulate gyri which are among top-10 values indicate their relations to excessive fluctuations of FC in DMN-related regions \cite{2020zdxx1}. Other top-10 regions are consistent with prior findings regarding identification of salient brain regions \cite{2021zdxx2, 2023zdxx3}.

\begin{figure}[htbp]
    \centering
    \includegraphics[width=0.96\linewidth]{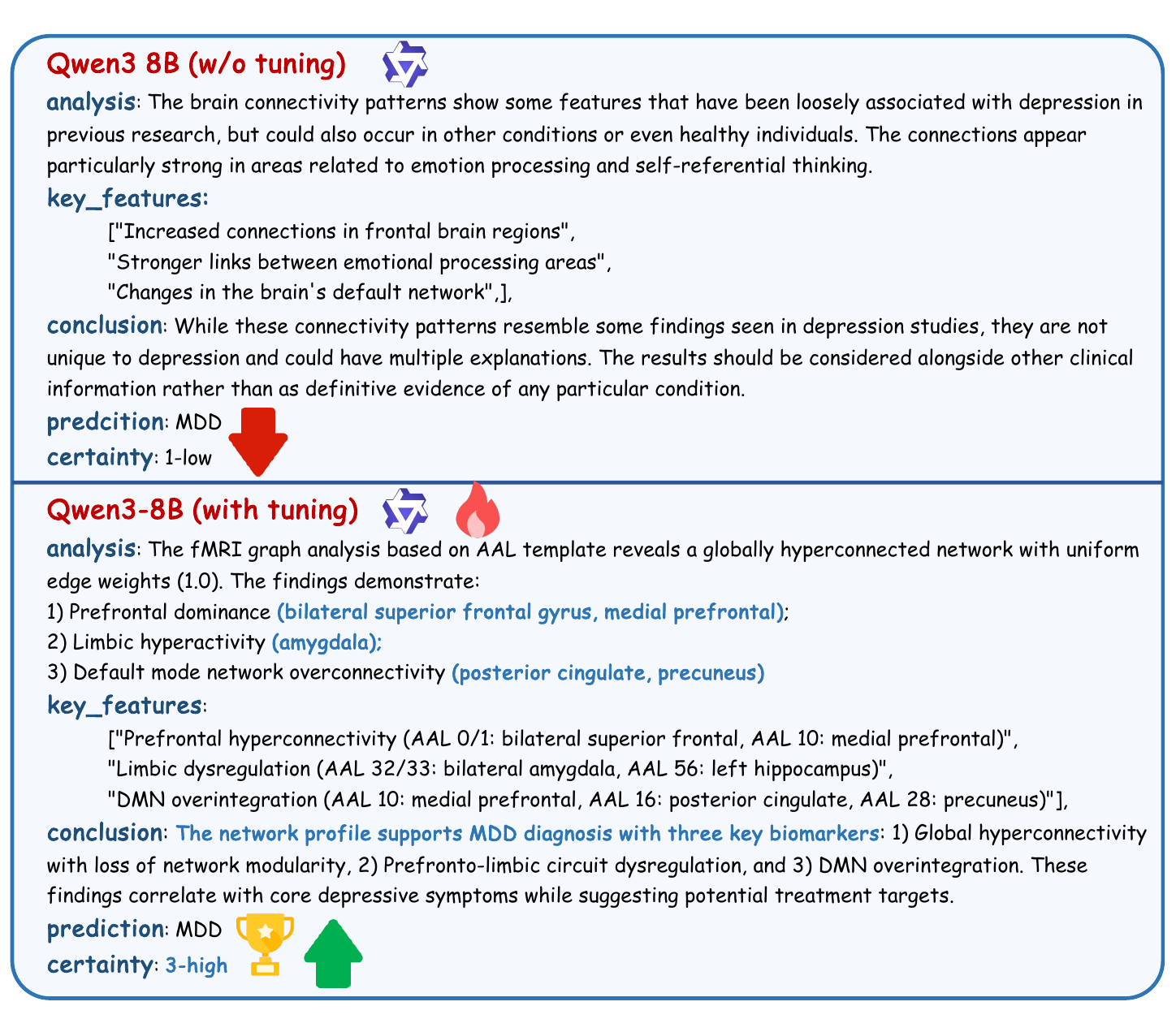}
    \caption{Text generation for ZDXX dataset from LM.}
    \label{fig_response}
\end{figure}
\vspace{-1em}

\subsection{Text generation for LM}
Here we mainly utilize LLM as enhancer which functions in embedding level. Yet we also select Qwen3-8B for tuning which has great capability of instruction following. The aim is to explore possibility of interpretable brain network analysis. As is shown in Fig. \ref{fig_response}, compared to vanilla Qwen, tuned LLM can generate more professional analysis on unseen private data, with more confidence for brain disease diagnosis judgment (MDD). It suggests a promising future of BLEG for further explainable and generalizable brain network analysis. Modern LM have the ability to capture deeper domain-specific knowledge from public datasets which can lead to better performance on private dataset.

\section{Conclusion}
\label{sec_conclusion}


In this work, we propose BLEG, a novel method that functions LLM as a powerful enhancer to boost GNN's performance in brain network analysis. Instead of directly training LLMs, we adopt a LLM-LM paradigm to leverage enhanced textual representations more efficiently. LM-aided SFT further enhances GNN's capability for downstream tasks. Extensive experiments on five datasets demonstrate effectiveness of BLEG, including different few-shot experiments which confirmed BLEG's great generalization. Our BLEG is a first attempt to leverage LLM with GNN-based brain network analysis and we think it can provide new insight for both research and real-world medical diagnose applications.

\section*{Acknowledgement}
This work is supported by National Natural Science Foundation of China (Grant No.62471133). 
This work is also supported by the Big Data Computing Center of Southeast University.


\appendix
\label{appendix}

\section{Related Works}
In this section we give a detailed description on the baselines in our comparison experiments.

\begin{itemize}
    \item \textbf{GCN} \cite{2017GCN}: Based on message passing mechanism, GCN aggregates the neighborhoods and then updates the value of the node.
    \item \textbf{GAT} \cite{2018GAT}: Via adding Attention to nodes, GAT updates the value by calculating attention scores of the neighborhood nodes. 
    \item \textbf{GraphSAGE} \cite{2017GraphSAGE}: GraphSAGE samples multi-hop neighborhood nodes and update its embedding, which is efficient in inductive training.

    \item \textbf{GraphTrans} \cite{wu2021representing}: GraphTrans is a transformer-based GNN method for graph-level tasks. It uses learnable GNN encoders before transformer layer to capture structure information.

    \item \textbf{BrainGNN} \cite{2021BrainGNN}: BrainGNN is a graph neural network method specifically designed for capturing the functional connectivity patterns between brain regions.

    \item \textbf{IBGNN} \cite{2022IBGNN}: IBGNN is an interpretable framework to analyze disorder-specific Regions of Interest (ROIs) and prominent connections. 

    \item \textbf{BrainNPT} \cite{2024BrainNPT} BrainNPT is a pre-trained transformer-based GNN method that learns general brain graph feature representations through pre-training. 

    \item \textbf{THFCN} \cite{2024thfcn}: THFCN enhances peroformance of functional connectivity networks by incorporating high-order features through hypergraph-based manifold regularization.

    \item \textbf{ContrastPool} \cite{2024ContrastPool}: ContrastPool is the latesst contrastive graph pooling method designed for interpretable classification of brain networks.

\end{itemize}

\section{Prompt Examples}
\label{append_prompt}

\subsection{Prompt Example for Augmented Text Generation}
\begin{figure*}[!h]
    \centering
    \includegraphics[width=\textwidth]{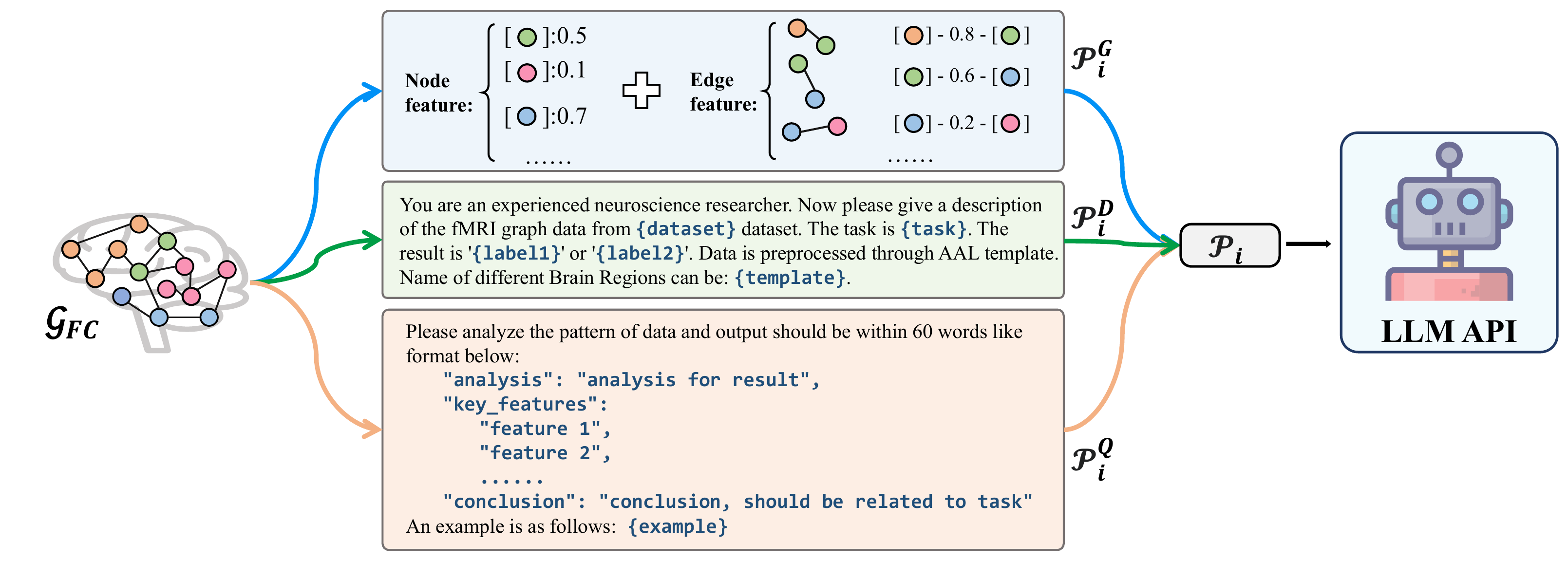}
    \caption{Prompt design for given FC graph.}
    \label{fig_prompt}
\end{figure*}

Here we provide an example for input FC brain network $\mathcal{G}_i$. Corresponding $\mathcal{P}_i^D$, $\mathcal{P}_i^G$, $\mathcal{P}_i^Q$ are shown in Fig. \ref{fig_prompt}.
We also list following response from LLM. We take one graph from ABIDE dataset just as an example (shown in Fig. \ref{fig_response}).

\subsection{Prompt Example for Tuned LM (Qwen3-8B)}
In this work we mainly utilize LLM as enhancer which functions in embedding level to assist GNNs. Yet we also conduct instruction tuning on Qwen3-8B which has great capability of instruction following. The aim is to explore possibility of interpretable brain network analysis through brain network based instruction tuning. Prompt design of this part in shown in Append. \ref{prompt_lm} and results are listed in Experiments part.

\section{Details for Datasets}
\label{append_dataset}

\begin{figure*}[!h]
    \centering
    \includegraphics[width=\textwidth]{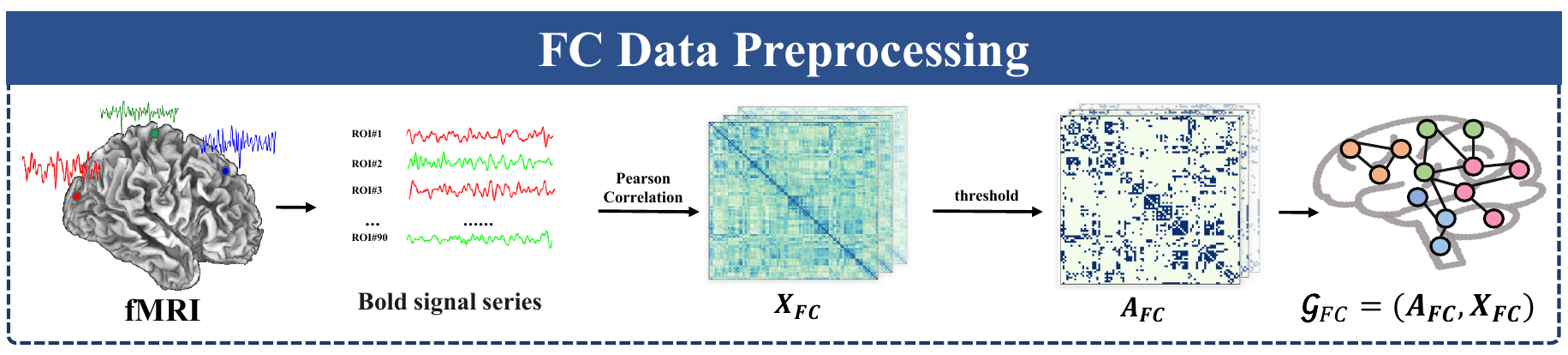}
    \caption{Preprocess of fMRI data and construction for FC dataset.}
    \label{fig_preprocess}
\end{figure*}

\textbf{fMRI preprocessing and dataset construction.}
Preprocess for fMRI data in shown in Fig. \ref{fig_preprocess}. The Data Preprocessing Assistant for Resting-State Function (DPARSF) MRI tookit \cite{DPARSF} is utilized for fMRI preprocessing. Then the average time series are computed for each brain region with AAL template. Pearson correlation is then calculated as function matrix, which denotes the feature matrix for FC ($X_{FC}$). Its adjacency matrix ($A_{FC}$) is obtained by thresholding a certain proportional quantization on the function matrix.

\begin{figure*}[!h]
    \centering
    \includegraphics[width=\textwidth]{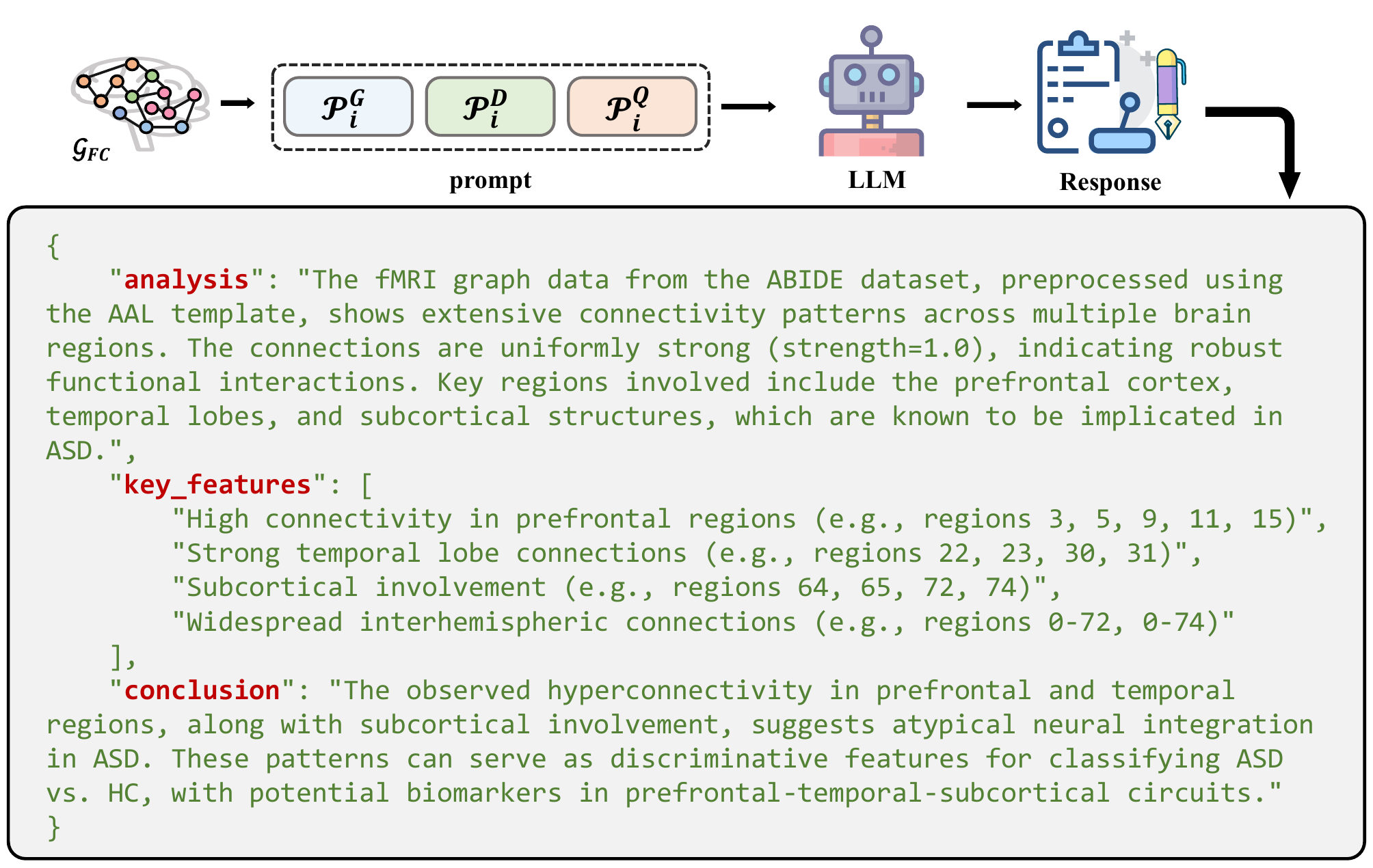}
    \caption{An example for prompt response from LLM of ABIDE dataset.}
    \label{fig_response}
\end{figure*}

\noindent\textbf{Textual dataset details.}
Total sample of textual datasets is equal to total number of public datasets in Tab. \ref{tab_append_dataset} which is 4760. The average length of input is 147.8 with output response length 103.4. For LLM generation evaluation and refinement, we follow practice from Qwen3 \cite{qwen3} and prompt LLM as judge to refine outputs from different dimensions. Note that evaluation for LLM output is still a opening yet challenging problem, especially in medical domain where accuracy of output is of extreme importance. Our results prove that LLM output can have positive impact on GNNs from latent representation level and validation for LLM generation are left as future work.

\noindent\textbf{More dataset details.} Details for datasets can be found in Tab. \ref{tab_append_dataset}. ABIDE dataset is for Autism Spectrum Disorder diagnosis (ASD). ADHD is a public dataset which focuses on Attention Deficit and Hyperactive Disorder disease (ADHD). HCP dataset is for gender classification. Rest-meta-MDD and zhongdaxinxiang datasets deal with Major Depressive Disorder diagnosis (MDD). HC in Tab. \ref{tab_append_dataset} means controls compared to patients (ASD, MDD, ADHD).

\begin{table*}[!htbp]
\centering
\caption{More details of datasets.}

\begin{tabular}{l|ccccc}
\toprule
\textbf{Datasets} & Tasks & Samples & Nodes ($\mathcal{|V|}$) & Classes &  Categories     
\\ \midrule
\textbf{ABIDE}       & ASD diagnosis      & 618 & 90      & 2 & \{HC, ASD\} \\
\textbf{ADHD}     & ADHD diagnosis     & 938 & 90     & 2 & \{HC, ADHD\} \\
\textbf{HCP}   & Gender classification      & 1039 & 90      & 2 & \{Male, Female\} \\
\textbf{Rest-meta-MDD}  & MDD diagnosis     & 2165 & 90       & 2 & \{HC, MDD\} \\
\midrule
\textbf{zhongdaxinxiang}      & MDD diagnosis      & 520 & 90       & 2 & \{HC, MDD\} \\
\bottomrule
\end{tabular}
\label{tab_append_dataset}
\end{table*}

\section{Details for Experimental Settings}
\label{append_exp}
More training details in instruction tuning stage can be found in Tab. \ref{tab_append_it}, including training arguments, LM settings and GCN encoder settings.

\begin{table*}[!htbp]
\begin{center}
\begin{tabular}{c|l|c}
\toprule
\textbf{Types} & \textbf{Parameters} & \textbf{Values} \\ 

\midrule
\multirow{4}{*}{Training Arguments} & 
Epochs                  &   5 \\ &
Dataset Length          &   4760 \\ &
Batch Size              &   32 \\ &
Learning Rate           &   $5 \times 10^{-5}$ \\

\midrule
\multirow{3}{*}{LM Settings} &
Model                   & BioGPT \\ &
Parameters              & 347M (1.57GB) \\ &
Hidden                  & 1024 \\

\midrule
\multirow{4}{*}{GCN Settings} &
Layers                  & 3 \\ &
Norm                    & \verb|BatchNorm()| \\ &
Activate                & \verb|GeLU()| \\ & 
Dropout                 & 0.3 \\

\bottomrule
\end{tabular}
\end{center}
\caption{Hyper-parameter settings for instruct-tuning.}
\label{tab_append_it}
\end{table*}

More training details in supervised fine-tuning stage can be found in Tab. \ref{tab_append_sft}, including training arguments, adapter settings and hyper-parameter searching space.

\begin{table*}[!htbp]
\begin{center}
\begin{tabular}{c|l|c}
\toprule
\textbf{Types} & \textbf{Parameters} & \textbf{Values} \\ \midrule

\multirow{5}{*}{Training Arguments} & 
Epochs                  &   150 \\ &
Early Stop              &   50 \\ &
Batch Size              &   $[32, 64, 128]$ \\ &
Learning Rate           &   $[0.0005, 0.0001]$ \\ &
Weight Decay            &   $[0, 0.0001]$ \\ &
LoRA rank               &   64 \\

\midrule
\multirow{4}{*}{Adapter Settings} &
Layers                  &   2 \\ &
Norm                    &   \verb|BatchNorm()| \\ &
Activate                &   \verb|GeLU()| \\ &
Dropout                 &   $[0.1, 0.2, 0.3, 0.4, 0.5]$ \\

\midrule
\multirow{3}{*}{Hyper-parameter Settings} &
Alignment Function      &   \verb|MSELoss()| \\ &
$\alpha$                &   $[0.0, 0.2, 0.3, 0.4, 0.5, 0.6, 0.7]$ \\ &
Few-shot ratio          &   $[0.1, 0.2, 0.3, 0.4, 0.5, 0.6, 0.7]$ \\

\bottomrule
\end{tabular}
\end{center}
\caption{Hyper-parameter settings for supervised fine-tuning.}
\label{tab_append_sft}
\end{table*}

\section{Proof of Theorem}
\begin{definition}
    \label{def_1}
    For original GNN, we define its representation as $\textbf{X}^{\mathcal{G}}$. For tuning LM, its representation is denoted as $\textbf{X}^{\mathcal{T}}$. Similarly, for the embeddings of BLEG where a distillation loss is added between two embedding, we define LM-aided GNN representation as $\textbf{X}^{\mathcal{G'}}$. And for given downstream task, its label information is denoted as $Y$. 
\end{definition}

\begin{assumption}
    \label{assum_1} 
    We leverage text information by finetuning LM based on LLM. For downstream label representation $Y$ and LM representation $\textbf{X}^{\mathcal{T}}$, we use mutual information to describe correlations between two representations. For $\textbf{X}^{\mathcal{G}}$, we assume $I(\textbf{X}^{\mathcal{T}};Y|\textbf{X}^{\mathcal{G}}) > 0$. 
\end{assumption}

\begin{assumption}
    \label{assum_2}
    Here we use $\mathcal{L}_{CE}$ and $\mathcal{L}_{align}$ respectively to optimize the model. And we assume following error bounds between different logits: $\mathbb{E}(||\textbf{X}^{\mathcal{G'}} - \textbf{X}^{\mathcal{T}}||^2) \le  \delta_1^2$, $\mathbb{E}(||\textbf{X}^{\mathcal{G}} - \textbf{X}^{\mathcal{G'}}||^2) \le \delta_2^2$, where $\delta_1, \delta_2 > 0$.

\end{assumption}

The first assumption states that LM contains complementary information of GNN that is useful for downstream tasks. This assumption is guaranteed by LLM's powerful capability of representation. Moreover, finetuning a smaller LM can also capture these useful text representation, which is already proved in TAPE \cite{2023tape}.
The second assumption states that via loss function as constraints, $(\textbf{X}^{\mathcal{G}}, \textbf{X}^{\mathcal{G'}})$, $(\textbf{X}^{\mathcal{G'}}, \textbf{X}^{\mathcal{T}})$ are aligned within certain error bounds.

\begin{theorem}
    \label{theory_1}
    \textbf{Complementary Representations from LM for GNN.}
    We define representation from original GNN ($\textbf{X}^{\mathcal{G}}$) and fine-tuned LM ($\textbf{X}^{\mathcal{T}}$), LM-distilled GNN representation is denoted as $\textbf{X}^{\mathcal{G'}}$. Downstream label representation is denoted as Y. Given above assumptions, we have: $\left\| I(\textbf{X}^{\mathcal{G'}};Y) -         I(\textbf{X}^{\mathcal{G}},\textbf{X}^{\mathcal{T}};Y)\right\| \le C \cdot \epsilon$, where $C$ is a constant and $\epsilon > 0$. Thus for $I(\textbf{X}^{\mathcal{G'}};Y)$ and $I(\textbf{X}^{\mathcal{G}};Y)$, $I(\textbf{X}^{\mathcal{G'}};Y) >  I(\textbf{X}^{\mathcal{G}};Y)$.

\end{theorem}

\label{append_proof}
Considering Assumption. \ref{assum_2}, and for model denoted as $f_\theta$, $P_\theta(Y|X)$ prs with respect to 
$X$. Then we have:

\begin{equation}
   \sup_Y
   \left\|
   P_{\theta}(Y|X_1)-P_{\theta}(Y|X_2)
   \right\|
   \leq 
   L \cdot
   \left\|
   X_1-X_2
   \right\| 
\end{equation}

Then here for $\XGraw$, $\XGnew$, we have:

\begin{small}
\begin{equation}
\begin{aligned}
        I(\textbf{X}^{\mathcal{G}'};Y)-I(\textbf{X}^{\mathcal{G}};Y)=\mathbb{E}_{\textbf{X}^{\mathcal{G}'}}
    \left[
    D_{\mathrm{KL}}(P(Y|\textbf{X}^{\mathcal{G}'})\|P(Y))
    \right]
    \\
    \quad \quad
    -\mathbb{E}_{\textbf{X}^{\mathcal{G}}}
    \left[
    D_{\mathrm{KL}}(P(Y|\textbf{X}^{\mathcal{G}})\|P(Y))
    \right]
\end{aligned}
\end{equation}
\end{small}

Again with Assumption \ref{assum_2}, we derive upper bound for different mutual information:

\begin{equation}
\begin{cases}
    \left\|
    I(\textbf{X}^{\mathcal{G}'};Y)-I(\textbf{X}^{\mathcal{G}};Y)
    \right\|
    \leq C_2\cdot\delta_2
    \\   
    I(\textbf{X}^{\mathcal{T}};Y|\textbf{X}^{\mathcal{G}'})\leq C_1\cdot\delta_1
\end{cases}
\label{equ_proof_1}
\end{equation}

Where $C_1$, $C_2$ are two constants.

For $I(\XGraw, \XTraw; Y)$, we have:

\begin{equation}
    I(\XGraw,\XTraw; Y) = I(\XGraw;Y) + I(\XTraw;Y|\XGraw)
\end{equation}

Thus we have:

\begin{small}
\begin{equation}
\begin{aligned}
    I(\XGraw,\XTraw;Y)-I(\XGnew;Y) =
    &
    \left[
    I(\XGraw;Y)-I(\XGnew;Y)
    \right] 
    \\& \quad
    + I(\XTraw;Y|\XGraw)
    \label{equ_proof_2}
\end{aligned}
\end{equation}
\end{small}

Then according to Eq. \ref{equ_proof_1}, we have:
\begin{equation}
    -C_2\cdot\delta_2\leq I(\XGraw;Y)-I(\XGnew;Y)\leq C_2\cdot\delta_2
    \label{equ_proof_3}
\end{equation}

Eq. \ref{equ_proof_2} can be written as:

\begin{equation}
\begin{aligned}
    &\left\|
    I(\XGraw, \XTraw; Y) - I(\XGnew;Y)
    \right\|
    \leq
    \\ &
    \quad \quad \quad \quad
    \left\|
    I(\XGraw; Y) - I(\XGnew; Y)
    \right\|
    + I(\XTraw; Y|\XGraw)
    \label{equ_proof_4}
\end{aligned}
\end{equation}

Combining Eq. \ref{equ_proof_1}, \ref{equ_proof_3} and \ref{equ_proof_4}, we can derive:

\begin{small}
\begin{equation}
\begin{aligned}
    -C_2\cdot\delta_2 \le
    I(\XGraw, \XTraw; Y) - I(\XGnew; Y)
    &\leq 
    C_1\cdot\delta_1 + C_2\cdot \delta_2
    \\
    \Rightarrow 
    \left\|
    I(\XGraw, \XTraw; Y) - I(\XGnew; Y)
    \right\| 
    &\le 
    C_1\cdot\delta_1 + C_2\cdot\delta_2
\end{aligned}
\end{equation}
\end{small}

We denote $C\cdot \epsilon \leftarrow (C_1\cdot\delta_1+C_2\cdot\delta_2)$, then we have:

\begin{equation}
    \left\|
    I(\XGraw, \XTraw; Y)-I(\XGnew; Y)
    \right\|
    \leq C \cdot \epsilon
    \label{equ_proof_5}
\end{equation}

Further, according to Assumption \ref{assum_1} and definition of mutual information, we have:

\begin{equation}
    I(\XGraw, \XTraw; Y) = I(\XGraw; Y) + I(\XTraw; Y|\XGraw) > I(\XGraw;Y)
    \label{equ_proof_6}
\end{equation}

Finally by combining Eq. \ref{equ_proof_5} and Eq. \ref{equ_proof_6}, we have:

\begin{equation}
    I(\XGnew; Y) > I(\XGraw; Y)
\end{equation}

Which means that through LM as enhancer, GNN can capture complementary information, leading to bigger value for mutual information between embeddings and downstream tasks.

\begin{tcolorbox}[promptbox=interest_colframe/interest_colback,nofloat=true,title={Prompt of tuned LM (Qwen3-8B)}, label=box:suspect-prompt]
\label{prompt_lm}
\colorbox{green!20}{\textbf{\# Task Description}}\\
You are an experienced neuroscience researcher. Now please give a description of the fMRI graph data from {dataset} dataset. The task is {task}. The result is '\textcolor{BlueViolet}{\{label1\}}' or '\textcolor{BlueViolet}{\{label2\}}'. Data is preprocessed through AAL template.
Name of different Brain Regions can be: \textcolor{BlueViolet}{\{template\}}

\colorbox{blue!20}{\textbf{\# Requirements}}

1. \textbf{Input data introduction}: Raw input data whose format is \textcolor{BlueViolet}{\{Data introduction\}}\\
2. \textbf{Analysis requirement}: Your analysis should be accurate and every conclusion must be supported by direct proof from input data.

\colorbox{yellow!20}{\textbf{\# Output Format}}
Your output should strict obey a json data whose structure is as follows:

\begin{verbatim}
{
\end{verbatim}
\hspace*{2em}``analysis": ``analysis for result",\\
\hspace*{2em}``key\_features":\\

\hspace*{4em}``feature 1",\\
\hspace*{4em}``feature 2",\\
\hspace*{4em}...\\
\hspace*{2em}``prediction": ``prediction of the data, must be aligned with task type",\\
\hspace*{2em}``certainty": ``Confidence of your prediction, a value between [1, 5]"
\begin{verbatim}
}
\end{verbatim}

\colorbox{Maroon!20}{\textbf{\# Input Data}}

- raw data:
\textcolor{BlueViolet}{\textbf{ \{Data\}}}\\
\end{tcolorbox}

\section{Broader Impacts}
\label{append_impact}
\subsection{Ethical Statements}
For private dataset zhongdaxinxiang, it is from Affiliated ZhongDa Hospital of Southeast University and the Second Affiliated Hospital of Xinxiang Medical University. 245 patients with a diagnosis of MDD and 275 age, gender and education level-matched healthy controls (HC) were recruited. All the participants completed a semi-structured clinical interview for DSM-IV Axis Disorders(SCID-I/P), clinician Version with two senior psychiatrists. They also had an identical assessment protocol,including review of medical history and demographic inventory.

Further, the research protocol was approved by the institutional ethics committee, and all participants provided written informed consent. To safeguard privacy, both raw and processed data are stored exclusively on internal servers and we use them just for researh but not practical deployment. We did not use any LLM APIs to direct analysis or transmit the data. Moreover, all subject identifiers were irreversibly removed, research personnel themselves have no access to these identity records, thereby precluding any possibility of personal information leakage.

The original MRI and questionnaire data are not publicly available due to privacy or confidentiality restrictions. The code used for the analyses is available in the supplementary materials. All data are available upon reasonable request from the corresponding author.

For public datasets, we strictly adhered to their respective usage agreements. All preprocessing pipelines follow official procedures provided by the dataset maintainers, and every evaluation metric employed in our study is fully aligned with the official benchmark settings.

\subsection{More Dataset Discussions}
We discuss more about possible negative social impacts. As part of the research in this paper deals with the diagnosis of depression (on a real-world private dataset), it is necessary to elaborate here on the possible negative social impacts of this work, despite the fact that all the current work is at the stage of scientific research and has not been put to practical use. Including but not limited to: 
\begin{itemize}
    \item Incorrect diagnosis. AI methods must have the possibility of error, which cannot be avoided, but an incorrect diagnosis will have a significant impact on individuals and society. Therefore, AI tools can only be used as a diagnostic aid, not as a decision maker, and the final decision should still be made by the doctor.

    \item Leakage of privacy information. In depression dataset, the identity information of the subjects is highly private, and the leakage of identity information will also have unpredictable and significant impact on individuals and society. Therefore, in this work, we have completely hidden the subjects’ identifying information (which is also not visible to the staff in the study group) as a way of preventing the leakage of private information.

    \item Role of BLEG in real-world diagnosis. Finally we position our method as an assistant rather than direct decision maker for doctors. We fully acknowledge that many of these complexities must be addressed before real-world deployment; however, covering every contingency is, at present, beyond the reach of any single AI approach. We therefore contend that a more effective and safer strategy is to treat our AI diagnostic model as an auxiliary instrument. In practice, doctors can inject domain-specific clinical knowledge such as regional epidemiology or demographic traits which requires little extra human labol (e.g. minor dataset adjustments or prompt tuning of the LLM) to achieve scenario-adaptive diagnostic results. These outputs then inform, rather than override, doctors' final decision. In short, while our work makes a constructive exploration in improving downstream performance and interpretability, it is ultimately designed to relieve human's labol and assist rather than replace human medical judgment.
\end{itemize}

\section{Future Works}
\label{append_future}
Although our novel attempt to enhance brain GNNs with LLMs has demonstrated promising results, BLEG still have space for improvement. Prompt design is crucial in fully activating strengths of LLMs and different LLM can have different generations. Efficient Training for LM can also influence final performance. As our BLEG is a general pipeline, we conclude our future work as follows: (a) More powerful LLM selection and GNN design. (b) More efficient prompt design and textual dataset generation, as well as generated data verification and refinement. (c) Other distillation strategies for better representations.

{
    \small
    \bibliographystyle{ieeenat_fullname}
    \bibliography{references}
}


\end{document}